\documentclass[10pt,twocolumn,letterpaper]{article}

\usepackage{cvpr}
\usepackage{times}
\usepackage{epsfig}
\usepackage{graphicx}
\usepackage{amsmath}
\usepackage{amssymb}
\usepackage[numbers]{natbib}
\usepackage[export]{adjustbox}

\usepackage[utf8]{inputenc} %
\usepackage[T1]{fontenc}    %
\usepackage{booktabs}       %
\usepackage{amsfonts}       %
\usepackage{nicefrac}       %
\usepackage{microtype}      %
\usepackage{mleftright,xparse}
\usepackage{enumitem}
\usepackage{footnote}
\makesavenoteenv{tabular}
\usepackage{fp}
\usepackage{xr}

\usepackage{color,xcolor}
\usepackage{epsfig}
\usepackage{graphicx}

\usepackage{adjustbox}
\usepackage{array}
\usepackage{booktabs}
\usepackage{colortbl}
\usepackage{float,wrapfig}
\usepackage{hhline}
\usepackage{array,multirow}
\usepackage{subcaption} %

\usepackage{amsmath,amsfonts,amsthm,amssymb}

\usepackage{bm}
\usepackage{nicefrac}
\usepackage{microtype}
\usepackage{dsfont}
\usepackage{amsmath}

\usepackage{changepage}
\usepackage{extramarks}
\usepackage{fancyhdr}
\usepackage{lastpage}
\usepackage{setspace}
\usepackage{soul}
\usepackage{xspace}
\usepackage{makecell}
\usepackage{mfirstuc}
\usepackage[pagebackref=true,breaklinks=true,letterpaper=true,colorlinks,bookmarks=false]{hyperref}

\usepackage{url}

\usepackage{algorithm}
\usepackage{algpseudocode}
\usepackage{capt-of}
\usepackage{enumerate}

\usepackage{caption}
\DeclareCaptionFont{ninept}{\fontsize{9pt}{11pt}\selectfont #1}
\usepackage[title]{appendix}

\usepackage[breaklinks=true,bookmarks=false]{hyperref}

\cvprfinalcopy %

\def\cvprPaperID{9906} %

\ifcvprfinal\fi

\newcommand{\discj}{D}

\newcommand{\gencj}{G}

\newcolumntype{L}[1]{>{\raggedright\let\newline\\\arraybackslash\hspace{0pt}}m{#1}}
\newcolumntype{C}[1]{>{\centering\let\newline\\\arraybackslash\hspace{0pt}}m{#1}}
\newcolumntype{R}[1]{>{\raggedleft\let\newline\\\arraybackslash\hspace{0pt}}m{#1}}

\newcommand{\hide}[1]{}

\newcommand{\reffig}[1]{Figure~\ref{fig:#1}}
\newcommand{\refsec}[1]{Section~\ref{sec:#1}}
\newcommand{\refapp}[1]{Appendix~\ref{sec:#1}}

\newcommand{\refeq}[1]{Equation~\eqref{eq:#1}}

\newcommand{\lblfig}[1]{\label{fig:#1}}
\newcommand{\lblsec}[1]{\label{sec:#1}}
\newcommand{\lbleq}[1]{\label{eq:#1}}

\newcommand{\ignorethis}[1]{}

\newcommand{\interior}[1]{%
  {\kern0pt#1}^{\mathrm{o}}%
}

\makeatletter
\newcommand\fs@spaceruled{\def\@fs@cfont{\bfseries}\let\@fs@capt\floatc@ruled
  \def\@fs@pre{\vspace{-2\baselineskip}\hrule height.8pt depth0pt \kern2pt}%
  \def\@fs@post{\kern2pt\hrule\relax}%
  \def\@fs@mid{\kern2pt\hrule\kern2pt}%
  \let\@fs@iftopcapt\iftrue}
\makeatother

\algrenewcommand{\algorithmicrequire}{\textbf{Input:}}
\algrenewcommand{\algorithmicensure}{\textbf{Output:}}
\let\originalleft\left
\let\originalright\right
\renewcommand{\left}{\mathopen{}\mathclose\bgroup\originalleft}
\renewcommand{\right}{\aftergroup\egroup\originalright}

\DeclareMathOperator*{\E}{\mathds{E}}

\newcommand{\pascalvoc}[1]{PASCAL-VOC\xspace}
\newcommand{\pascalvocshort}[1]{PASCAL-VOC\xspace}

\newcommand{\ignore}[1]{}

\newcommand{\fid}{Fr\'echet Inception Distance\xspace}

\def\rvr{{\mathbf{r}}}

\def\rvx{{\mathbf{x}}}

\def\rvz{{\mathbf{z}}}

\makeatletter
\DeclareRobustCommand\onedot{\futurelet\@let@token\@onedot}
\def\@onedot{\ifx\@let@token.\else.\null\fi\xspace}

\def\eg{e.g\onedot}

\makeatother

\definecolor{MyDarkBlue}{rgb}{0,0.08,1}
\definecolor{MyDarkGreen}{rgb}{0.02,0.6,0.02}
\definecolor{MyDarkRed}{rgb}{0.8,0.02,0.02}
\definecolor{MyDarkOrange}{rgb}{0.40,0.2,0.02}
\definecolor{MyPurple}{RGB}{111,0,255}
\definecolor{MyRed}{rgb}{1.0,0.0,0.0}
\definecolor{MyGold}{rgb}{0.75,0.6,0.12}
\definecolor{MyDarkgray}{rgb}{0.66, 0.66, 0.66}

\setlist[itemize]{leftmargin=8mm}
\newcommand{\myparagraphfirst}[1]{\paragraph{#1}}
\newcommand{\myparagraph}[1]{\vspace{-5pt}\paragraph{#1}}

\theoremstyle{definition}

\theoremstyle{plain}

\NewDocumentCommand\xDeclarePairedDelimiter{mmm}
{%
    \NewDocumentCommand#1{som}{%
    \IfNoValueTF{##2}
        {\IfBooleanTF{##1}{#2##3#3}{\mleft#2##3\mright#3}}
        {\mathopen{##2#2}##3\mathclose{##2#3}}%
    }%
}
\xDeclarePairedDelimiter{\size}{\lvert}{\rvert}
\xDeclarePairedDelimiter{\abs}{\lvert}{\rvert}
\xDeclarePairedDelimiter{\norm}{\lVert}{\rVert}
\xDeclarePairedDelimiter{\ceil}{\lceil}{\rceil}
\xDeclarePairedDelimiter{\floor}{\lfloor}{\rfloor}

\newcommand{\mhide}[1]{#1}

\newcommand{\new}{\textcolor{blue}}
\renewcommand{\new}{}

\title{Diverse Image Generation via Self-Conditioned GANs}

\author{Steven Liu\textsuperscript{1} \quad Tongzhou Wang\textsuperscript{1} \quad David Bau\textsuperscript{1} \quad Jun-Yan Zhu\textsuperscript{2} \quad Antonio Torralba\textsuperscript{1}
\\
\textsuperscript{1}MIT CSAIL \qquad \textsuperscript{2}Adobe Research}

\begin{document}

\twocolumn[{%
\renewcommand\twocolumn[1][]{#1}%
\maketitle
  \centering
    \vspace{-.15in}
  \includegraphics[width=\linewidth]{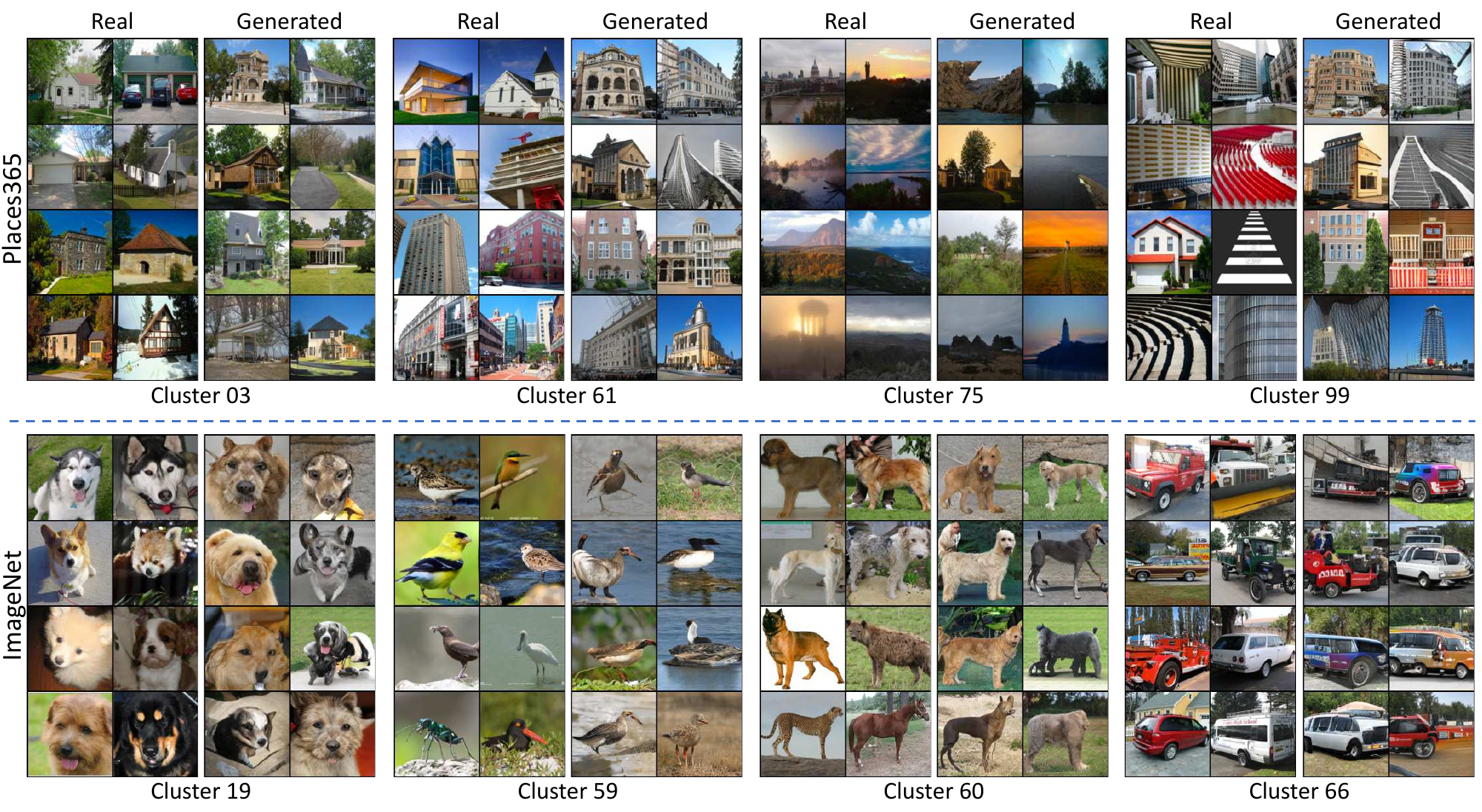}
  \vspace{-.3in}
  \captionof{figure}{Our proposed self-condition\new{ed} GAN model learns to perform clustering and image synthesis simultaneously. The model training requires no manual annotation of object classes. Here, we visualize several discovered clusters for both Places365 (top) and ImageNet (bottom). For each cluster, we show both real images and the generated samples conditioned on the cluster index. }
  \vspace{.1in}\vspace{1.5em}
  \lblfig{teaser}
}]
\maketitle

\begin{abstract}
We introduce a simple but effective unsupervised method for generating realistic and diverse images. 
We train a class-conditional GAN model without using manually annotated class labels. Instead, our model is conditional on labels automatically derived from clustering in the discriminator's feature space. Our clustering step automatically discovers diverse modes, and explicitly requires the generator to cover them. 
Experiments on standard mode collapse benchmarks show that our method outperforms several competing methods when addressing mode collapse. Our method also performs well on large-scale datasets such as ImageNet and Places365, improving both image diversity and standard quality metrics, compared to previous methods.
\end{abstract}
\section{Introduction}

Despite the remarkable progress of Generative Adversarial Networks (GANs)~\citep{goodfellow2014generative,brock2019large,karras2019style}, there remains a significant gap regarding the quality and diversity between \emph{class-conditional} GANs trained on labeled data, and \emph{unconditional} GANs trained without any labels in a fully unsupervised setting~\citep{mirza2014conditional,lucic2019high}. This problem reflects the challenge of \textit{mode collapse}: the tendency for a generator to focus on a subset of modes to the exclusion of other parts of the target distribution~\citep{goodfellow2016nips}. Both empirical and theoretical studies have shown strong evidence that real data has a highly multi-modal distribution~\citep{narayanan2010sample,tenenbaum2000global}. Unconditional GANs trained on such data distributions often completely miss important modes, \eg, not being able to generate one of ten digits for MNIST~\citep{salimans2016improved}, or omitting object classes such as people and cars within synthesized scenes~\citep{bau2019seeing}. Class-conditional GANs alleviate this issue by enforcing labels that require the generator to cover all semantic categories. However, in practice, it is often expensive to obtain labels for large-scale datasets. 

In this work, %
we present a simple but effective training method, self-conditioned GANs, to address mode collapse. We train a class-conditional GAN and automatically obtain image classes by clustering in the discriminator's feature space. Our algorithm alternates between learning a feature representation for our clustering method and learning a better generative model that covers all the clusters.  %
Such partitioning automatically discovers modes the generator is currently missing, and explicitly requires the generator to cover them. \reffig{teaser} shows several discovered clusters and corresponding generated images for each cluster. %

Empirical experiments demonstrate that this approach successfully recovers modes on standard mode collapse benchmarks (mixtures of Gaussians, stacked MNIST, CIFAR-10).  More importantly, our approach scales well to large-scale image generation, achieving better \fid \citep{heusel2017gans}, \new{Fr\'{e}chet Segmentation Distance \citep{bau2019seeing},} and Inception Score \citep{salimans2016improved} for both ImageNet and Places365, compared to previous unsupervised methods. 

Our \href{https://github.com/stevliu/self-conditioned-gan}{code} and models are available on our \href{http://selfcondgan.csail.mit.edu/}{website}. 

\section{Related Work}
\lblfig{related}

\myparagraphfirst{Generative Adversarial Networks (GANs).} Since the introduction of GANs~\citep{goodfellow2014generative}, many variants have been proposed~\citep{mirza2014conditional,denton2015deep,radford2015unsupervised,salimans2016improved,arjovsky2017wasserstein,mao2017least,gulrajani2017improved,mescheder2018training}, improving both the training stability and image quality. Due to its rapid advance, GANs have been used in a wide range of computer vision and graphics applications~\citep{wang2016generative,isola2017image,zhang2017stackgan,zhu2017unpaired,huang2018multimodal,hoffman2018cycada}. GANs excel at synthesizing photorealistic images for a specific class of images such as faces and cars~\citep{karras2018progressive,karras2019style}.  However, for more complex datasets such as ImageNet, state-of-the-art models are class-conditional GANs that require ground truth image class labels during training~\citep{brock2019large}. To reduce the cost of manual annotation, a recent work~\citep{lucic2019high} presents a semi-supervised method based on RotNet~\cite{gidaris2018unsupervised}, a self-supervised image rotation feature learning method. The model is trained with labels provided on only a subset of images.  On the contrary, our general-purpose method is not image-specific, and fully unsupervised. \refsec{large-scale} shows that our method outperforms a RotNet-based baseline. A recent method~\cite{mukherjee2018clustergan} proposes to obtain good clustering using GANs, while we aim to achieve realistic and diverse generation.%

\myparagraph{Mode collapse.} Although GANs are formulated as a minimax game in which each generator is evaluated against a discriminator, during optimization, the generator faces a slowly-changing discriminator that can guide generation to collapse to the point that maximizes the discriminator~\citep{metz2017unrolledgan}. Mode collapse does occur in practice, and it is one of the fundamental challenges for GANs training~\citep{goodfellow2016nips,salimans2016improved, khayatkhoei2018disconnected}. 

Several solutions to mode collapse have been proposed, including amending the adversarial loss to look ahead several moves (e.g., Unrolled GAN~\citep{metz2017unrolledgan}), jointly training an encoder to recover latent distributions (e.g., VEEGAN~\citep{guyon2017veegan}), \new{adding an auxiliary loss to prevent the catastrophic forgetting~\citep{chen2019self}}, packing the discriminator with sets of points instead of singletons~\citep{salimans2016improved,lin2018pacgan} and training a mixture of generators with an auxiliary diversity objective~\citep{hoang2018mgan,ghosh2018multi}.  %
Different from the above work, our method partitions the real distribution instead of the generated distribution, and devotes a class-conditioned discriminator to each target partition. 

Another related line of research trains class-conditional GANs on unlabelled images by clustering on features obtained via unsupervised feature learning methods~\citep{lucic2019high,sage2018logo} or pre-trained classifiers~\cite{sage2018logo}. 
In contrast, our method directly clusters on discriminator features that inherently exist in GANs, leading to a simpler method and achieving higher quality generation in our experiments (\refsec{large-scale}). Mixture Density GAN proposes to use log-likelihoods of a Gaussian mixture distribution in discriminator feature space as the GAN objective~\citep{eghbal2019mixture}. GAN-Tree uses clustering to split a GAN into a tree hierarchy of GANs for better mode coverage~\citep{kundu2019gan}. These methods, while also using clustering or mixture models, are mostly orthogonal with our work. Furthermore, the simplicity of our method allows it to be easily combined with a variety of these techniques.

\begin{figure}[t]
\centering
\includegraphics[width=\columnwidth]{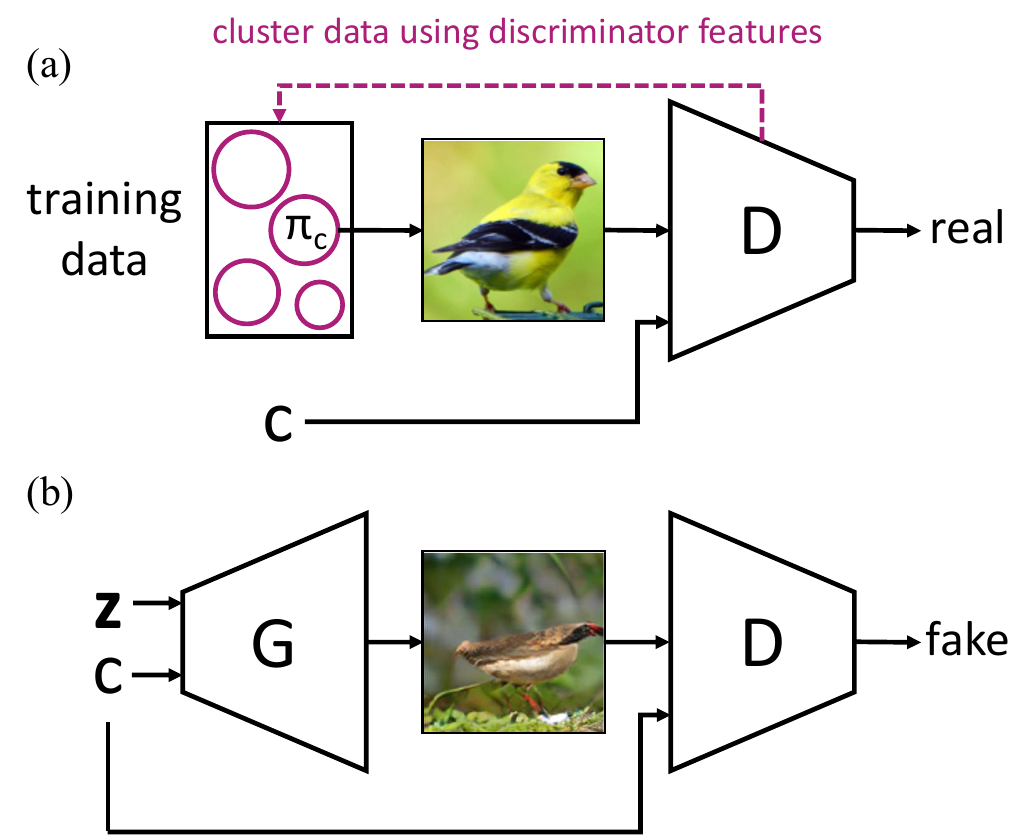}
\vspace{-14pt}
\caption{We learn a discriminator $D$ and a generator $G$ that are both conditioned on the automatically discovered cluster $c$.  %
\textbf{(a)} For a specific $c$, the discriminator $D$ must learn to recognize real images sampled from the cluster $\pi_c$ of the dataset, and distinguish them from (b) fake images synthesized by the class-conditional generator $G$. %
\textbf{(b)} The class-conditional generator $G$ synthesizes images from $\rvz$. By learning to fool $D$ when also conditioned on $c$, the generator learns to mimic the images in $\pi_c$. %
Our method differs from a conventional conditional GAN as we do not use ground-truth labels to determine the partition $\{\pi_c\}_{c=1}^k$.  Instead, our method begins with clustering random discriminator features and periodically reclusters the dataset based on discriminator features.}
\lblfig{architecture}%
\end{figure}
\section{Method}
\lblsec{method}
One of the core problems in generating diverse outputs in a high-dimen\-sional space such as images is mode collapse: the support of the generator's output distribution can be much smaller than the support of the real data distribution. One way mode collapse has been empirically lessened is by use of a class-conditional GAN, which explicitly penalizes the generator for not having support on each class.

We propose to exploit this class-conditional architecture, but instead of assuming access to true class labels, we will synthesize labels in an unsupervised way.
On a high level, our method dynamically partitions the real data space into different clusters, which are used to train a class-conditional GAN. Because generation conditioned on a cluster index is optimized with respect to the corresponding conditioned discriminator, and each discriminator is responsible for a different subset of the real distribution, our method encourages the generator output distribution to cover all partitions of the real data distribution.

Specifically, to train a GAN to imitate a target distribution $p_{\mathsf{real}}$, we partition the data set into $k$ clusters $\{\pi_1, \dots, \pi_k\}$ that are determined during training.  No ground-truth labels are used; training samples are initially clustered in the randomly initialized discriminator feature space,
and the clusters are updated periodically.  A class-conditional GAN structure is used to split the discriminator and the generator. Next, we describe two core components of our algorithm: \begin{itemize}
    \item Conditional GAN training with respect to cluster labels given by the current partitioning.
    \item Updating the partition according to the current discriminator features of real data periodically.
\end{itemize}
\vspace{-10pt}

\myparagraph{Conditional GAN training.}
\lblsec{cgan_training}
\begin{algorithm}
\caption{Self-Conditioned GAN Training}\label{algorithm}
\begin{algorithmic}
\State Initialize generator $\gencj$ and discriminator $\discj$
\State Partition dataset into $k$ sets $\{\pi_1, ... ,\pi_k\}$ using $D_f$ outputs
\For{number of training epochs}
\State {\textcolor[HTML]{727272}{// Conditional GAN training based on current partition}}
\For{number of training iterations for an epoch}
\For{$j$ in $\{1, ... , m\}$} 
\State Sample cluster $c^{(j)} \sim P_\pi$, where $c$ is chosen
\State \hspace{1.5em} with probability proportional to $|\pi_c|$.
\State Sample image $x^{(j)}\sim \pi_{c^{(j)}}$ from cluster $c^{(j)}$.
\State Sample latent $z^{(j)}\sim \mathcal{N}(0, I)$.
\EndFor
\State Update $\gencj$ and $\discj$ according to $\min_G \max_D \mathcal{L}_\mathsf{GAN}$  
\State \hspace{1.5em} on minibatch $\{(c^{(j)}, x^{(j)}, z^{(j)})\}_j$. \Comment{Eqn.~\eqref{eq:gan-minimax}}
\EndFor
\State {\textcolor[HTML]{727272}{// Clustering to obtain new partitions}}
\State Cluster on $D_f$ outputs of a subset of training set to 
\State \hspace{1.5em} identify a new partition $\{\pi^\mathsf{new}_c\}$ into $k$ sets, using
\State \hspace{1.5em} previous centroids as initialization. \Comment{Eqn.~\eqref{eq:recluster-init}}
\State Find the matching $\rho(\cdot)$ between $\{\pi^\mathsf{new}_c\}_c$ and $\{\pi_c\}_c$ 
\State \hspace{1.5em} that minimizes $\mathcal{L}_\mathsf{match}$. \Comment{Eqn.~\eqref{eq:match-loss}}
\State Update all $\pi_c \leftarrow \pi^\mathsf{new}_{\rho(c)}$.
\EndFor
\end{algorithmic}
\end{algorithm}
The GAN consists of a class-conditional generator $G(\rvz, c)$ associated with a class-conditional discriminator $D(\rvx, c)$.  We denote the internal discriminator feature layers as $D_f$ and its last layer as $D_h$ so $D = D_h \circ D_f$. The generator and discriminator are trained to optimize the following adversarial objective:
\begin{align}
\mathcal{L}_\mathsf{GAN}(D, G) = \E_{c\sim P_\pi} \bigg[& \E_{\rvx\sim \pi_c} \left[\log(D(\rvx,c)\right] 
 \nonumber\\
& +\E_{\rvz\sim \mathcal{N}(0, I)}\left[\log(1 - D(G(\rvz, c), c))\right]\bigg], \nonumber
\end{align}
where the cluster index $c$ is sampled from the categorical distribution $P_\pi$ that weights each cluster proportional to its true size in the training set. Here $G$ aims to generate images $G(z, c)$ that look similar to the real images of cluster $c$ for $z \sim \mathcal{N}(0, I)$, while $D(\cdot, c)$ tries to distinguish between such generated images and real images of cluster $c$. They are jointly optimized in the following minimax fashion:
$$    \min_G \max_D \mathcal{L}_\mathsf{GAN}(D, G). \label{eq:gan-minimax}$$
When under the condition $c$, the discriminator is encouraged to give low score for any sample that is not from cluster $c$ because $p_{\mathsf{real}}(\rvx \,|\, c) = 0$ for all $\rvx\notin \pi_c$.  So the corresponding conditioned generator is penalized for generating points that are not from cluster $c$, which ultimately prevents the generator from getting stuck on other clusters. The optimization is shown in \reffig{architecture}.  
\myparagraph{Computing new partition by clustering.}
As the training progresses, the shared discriminator layers $D_f$ learn better representations of the data, so we periodically update $\pi$ by re-partitioning the target dataset over a metric induced by the current discriminator features.  We use $k$-means clustering \citep{lloyd1982least} to obtain a new partition into $k$ clusters $\{\pi_c\}_{c=1}^k$ according to the $D_f$ output space, approximately optimizing
\begin{align}
\mathcal{L_{\mathsf{cluster}}}(\{\pi_c\}_{c=1}^k) = \E_{c\sim P_\pi} \left[ \E_{\rvx \sim \pi_c}  \left[ \norm{D_f(\rvx) - \boldsymbol{\mu}_c}_2^2 \right] \right],
\lbleq{kmeans}
\end{align}
where $\boldsymbol{\mu}_c \triangleq \frac{1}{\size{\pi_c}} \sum_{x \in \pi_c} D_f(\rvx)$ is the mean of each cluster in $D_f$ feature space. %

\myparagraph{Clustering initialization.} For the first clustering, we use the $k$-means++ initialization \citep{arthur2007kmeans++}. For subsequent reclustering, we initialize with the means induced by the previous clustering. That is, if $\{ \pi_c^\mathsf{old} \}_{c = 1}^{k}$ denotes the old cluster means and $\{ \boldsymbol{\mu}_c^{\mathsf{init}} \}_{c = 1}^{k}$ denotes the $k$-means initialization to compute the new clustering, we take  
\begin{equation}
    \boldsymbol{\mu}_c^{\mathsf{init}} = \frac{1}{\size{\pi_c^\mathsf{old}}} \sum_{x \in \pi_c^\mathsf{old}} D_f^\mathsf{new}(x), \qquad \forall c, \label{eq:recluster-init}
\end{equation}where $D_f^\mathsf{new}$ denotes the current discriminator feature space.%

\myparagraph{Matching with old clusters.}
After repartitioning, to avoid retraining the conditional generator and discriminator from scratch, we match the new clusters $\{\pi^\mathsf{new}_c\}_{c=1}^{k}$ to the old clusters $\{\pi^\mathsf{old}_c\}_{c=1}^{k}$ so that the target distribution for each generator does not change drastically. We formulate the task as a min-cost matching problem, where the cost of matching a $\pi^\mathsf{new}_c$ to a $\pi^\mathsf{old}_{c'}$ is taken as $\lvert \pi^\mathsf{old}_{c'} \setminus \pi^\mathsf{new}_c \rvert$, the number of samples missing in the new partition. Therefore, we aim to find a permutation $\rho \colon [k] \rightarrow [k]$ that minimizes the objective:
\begin{equation}
   \mathcal{L_{\mathsf{match}}}(\rho) = \sum_c \lvert \pi^\mathsf{old}_{c} \setminus \pi^\mathsf{new}_{\rho(c)}\rvert. \label{eq:match-loss}
\end{equation} For a given new partitioning from $k$-means, we solve this matching using the classic Hungarian min-cost matching algorithm \citep{kuhn1955hungarian}, and obtain the new clusters to be used for GAN training in future epochs. Algorithm~\ref{algorithm} summarizes the entire training method.

\myparagraph{Online clustering variant.} We have also experimented with online $k$-means based on gradient descent~\cite{bottou1995convergence}, where we updated the cluster centers and membership using \refeq{kmeans} in each iteration. Our online variant achieves comparable results on mode collapse benchmarks (\refsec{toy}), but performs worse for real image datasets (\refsec{mnist_cifar}), potentially due to the training instability caused by frequent clustering updates. Additionally, in \refsec{results}, we perform an ablation studies regarding clustering initialization, online vs.~batch clustering, and the clustering matching method.

\section{Experiments}
\lblsec{results}

\subsection{Experiment Details}

\paragraph{Network architecture.}
\new{To condition the generator, we learn a unit norm embedding vector for each class, which is fed with the latent input into the first layer of the generator. To condition the discriminator, we let the discriminator output a $k$-dimensional vector where $k$ is the number of clusters, and mask the output according to the input image's class label. Across all experiments, we use this method to add conditioning to unconditional backbone networks.}

For experiments on synthetic data, our unconditional generator and discriminator adopt the structure proposed in PacGAN~\citep{lin2018pacgan}. We use a $32$-dimensional embedding. 

For experiments on Stacked-MNIST, we use the DCGAN architecture~\citep{radford2015unsupervised}, following prior work \citep{lin2018pacgan}. For experiments on CIFAR-10 \citep{cifar10}, we use the DCGAN architecture, following SN-GANs \citep{miyato2018spectral}. For experiments on Places365~\citep{zhou2014learning} and ImageNet~\citep{deng2009imagenet}, we adopt the conditional architecture proposed by Mescheder \textit{et al.}~\cite{mescheder2018training}, and our unconditional GAN baseline removes all conditioning on the input label. For these four datasets, we use a $256$-dimensional embedding. 

\myparagraph{Clustering details.}
By default, we use $k=100$ clusters. We recluster every $10{,}000$ iterations for experiments on synthetic data, every $25{,}000$ iterations for Stacked-MNIST and CIFAR-10 experiments, and every $75{,}000$ iterations for ImageNet and Places365 experiments.  
The details of the online clustering variant are described in \refapp{app-details}.

\myparagraph{Training details.}
For experiments on synthetic datasets, CIFAR-10, and Stacked-MNIST, we train on the standard GAN loss proposed by Goodfellow \textit{et al.}~\cite{goodfellow2014generative}.  For experiments on Places365 and ImageNet, our loss function is the vanilla loss function proposed by Goodfellow \textit{et al.}~\cite{goodfellow2014generative} with the $R_1$ regularization as proposed by Mescheder \textit{et al.}~\cite{mescheder2018training}. \new{We find that our method works better with a small regularization weight $\gamma = 0.1$ instead of the default $\gamma = 10$. We explore this choice of hyperparameter in \refapp{app-reg}.}

\begin{table*}%
  \vspace{-10pt}
  \vspace{-8pt}
\centering%
  \caption{Number of modes recovered, percent high quality samples, and reverse KL divergence metrics for 2D-Ring and 2D-Grid experiments. \mhide{Results are averaged over ten trials, with standard error reported. }
  }
  \vspace{-2.5mm}
  \label{2d_results}
    \resizebox{%
        0.88\width
    }{!}
    {%
    \begin{tabular}{lrrlrrl}
      \toprule
      \multicolumn{1}{c}{} &
      \multicolumn{3}{c}{2D-Ring} &
      \multicolumn{3}{c}{2D-Grid} \\
      \cmidrule(r){2-4}
      \cmidrule(r){5-7}
              & \makecell[c]{Modes \\ (Max 8)}$~\uparrow$ & \%$~\uparrow$ & \multicolumn{1}{c}{Reverse KL$~\downarrow$}  & \makecell[c]{Modes \\ (Max 25)}$~\uparrow$    &  \%$~\uparrow$ & \multicolumn{1}{c}{Reverse KL$~\downarrow$}    \\
      \midrule
      
      GAN~\cite{goodfellow2014generative}
      & $6.3 \mhide{\pm 0.5}$ & $98.2 \mhide{\pm 0.2}$ & $0.45 \mhide{\pm   0.09}$ 
      & $17.3 \mhide{\pm 0.8}$ & $94.8 \mhide{\pm 0.7}$ & $0.70 \mhide{\pm 0.07}$ \\
      PacGAN2~\cite{lin2018pacgan}
      & $7.9 \mhide{\pm 0.1}$ & $95.6 \mhide{\pm 2.0}$ & $0.07 \mhide{\pm 0.03}$ 
      &   $23.8 \mhide{\pm 0.7}$ & $91.3 \mhide{\pm 0.8}$ & $0.13 \mhide{\pm 0.04}$ \\
      PacGAN3~\cite{lin2018pacgan} 
      & $7.8 \mhide{\pm 0.1}$ & $97.7 \mhide{\pm 0.3}$ & $0.10 \mhide{\pm 0.02}$ 
      &   $24.6 \mhide{\pm 0.4}$ & $94.2 \mhide{\pm 0.4}$ & $0.06 \mhide{\pm 0.02}$ \\
      PacGAN4~\cite{lin2018pacgan} 
      & $7.8 \mhide{\pm 0.1}$ & $95.9 \mhide{\pm 1.4}$ & $0.07 \mhide{\pm 0.02}$ 
      &   ${24.8} \mhide{\pm 0.2}$ & $93.6 \mhide{\pm 0.6}$ & $0.04 \mhide{\pm 0.01}$ \\
      Random Labels ($k=50$) 
      & $7.9 \mhide{\pm 0.1}$ & $96.3 \mhide{\pm 1.1}$ & $0.07 \mhide{\pm 0.02}$ 
      & $16.0 \mhide{\pm 1.0}$ & $90.6 \mhide{\pm 1.6}$ & $0.57 \mhide{\pm 0.07}$ \\
      MGAN ($k=50$)~\cite{hoang2018mgan} 
       & $\textbf{8.0} \mhide{\pm 0.0}$ & $71.7\mhide{\pm 1.3}$ & $0.0073\mhide{\pm 0.0014}$ 
       & $\textbf{25.0} \mhide{\pm 0.0}$ & $40.7\mhide{\pm 1.2}$ & $0.0473\mhide{\pm 0.0047}$ \\ 
    \new{Ours + Online Clustering ($k=50$) }
      & $\textbf{8.0} \mhide{\pm 0.0}$ & $\textbf{99.7} \mhide{\pm 0.1}$ & $\textbf{0.0014} \mhide{\pm 0.0001}$ 
      & $\textbf{25.0} \mhide{\pm 0.0}$ & $\textbf{99.7} \mhide{\pm 0.1}$  & $\textbf{0.0057} \mhide{\pm 0.0004}$ \\
      
      Ours ($k=50$) 
      & $\textbf{8.0} \mhide{\pm 0.0}$ & ${99.5} \mhide{\pm 0.3}$ & $\textbf{0.0014} \mhide{\pm 0.0002}$ 
      & $\textbf{25.0} \mhide{\pm 0.0}$ & ${99.5} \mhide{\pm 0.1}$  & ${0.0063} \mhide{\pm 0.0007}$ \\
      
      \bottomrule
    \end{tabular}%
  }%
  \vspace{1pt}
  \vspace{-1pt}%
\end{table*}
\begin{table*}%
  \vspace{-5pt}%
  \centering%
\caption{Number of modes recovered, reverse KL divergence, and Inception Score (IS) metrics for Stacked MNIST and CIFAR-10 experiments. \mhide{Results are averaged over five trials, with standard error reported.} Results of \hide{Unrolled GAN, VEEGAN and} PacGAN on Stacked MNIST are taken from \citep{lin2018pacgan}. For CIFAR-10, all methods recover all $10$ modes.}
\label{small_results}
  \vspace{-2.5mm}

\resizebox{%
\ifdim\width>\textwidth
  \textwidth
\else
  0.88\width
\fi
}{!}
{%
  \begin{tabular}{lrrrrr}
    \toprule
    \multicolumn{1}{c}{} &
    \multicolumn{2}{c}{Stacked MNIST} &
    \multicolumn{3}{c}{CIFAR-10} \\ 
    \cmidrule(r){2-3}
    \cmidrule(r){4-6}
            & \makecell{Modes \\ (Max 1000)}$~\uparrow$ & \multicolumn{1}{c}{Reverse KL$~\downarrow$}  & \multicolumn{1}{c}{FID$~\downarrow$}  & \multicolumn{1}{c}{IS$~\uparrow$}  & \multicolumn{1}{c}{Reverse KL$~\downarrow$}    \\
    \midrule
    GAN~\cite{goodfellow2014generative} & 
    $133.4\mhide{\pm 17.70}$ & $2.97 \mhide{\pm 0.216}$ & $28.08 \mhide{\pm 0.47}$ &  ${6.98} \mhide{\pm 0.062}$ & ${0.0150} \mhide{\pm 0.0026}$ \\
    PacGAN2~\cite{lin2018pacgan}  & 
    $\textbf{1000.0} \mhide{\pm 0.00}$ & $\textbf{0.06} \mhide{\pm 0.003}$ & $27.97 \mhide{\pm 0.63}$ &  ${7.12} \mhide{\pm 0.062}$ & ${0.0126} \mhide{\pm 0.0012}$ \\
    PacGAN3~\cite{lin2018pacgan}  & 
    $\textbf{1000.0} \mhide{\pm 0.00}$ & $\textbf{0.06} \mhide{\pm 0.003}$ &  $32.55 \mhide{\pm 0.92}$ &  ${6.77} \mhide{\pm 0.064}$ & ${0.0109} \mhide{\pm 0.0011}$ \\
    PacGAN4~\cite{lin2018pacgan} & 
    $\textbf{1000.0} \mhide{\pm 0.00}$ & $0.07 \mhide{\pm 0.005}$ & $34.16 \mhide{\pm 0.94}$ &  ${6.77} \mhide{\pm 0.079}$ & ${0.0150} \mhide{\pm 0.0005}$ \\
    Logo-GAN-AE~\cite{sage2018logo} & 
    $\textbf{1000.0} \mhide{\pm 0.00}$ & $0.09 \mhide{\pm 0.005}$ &  $32.49 \mhide{\pm 1.37}$ &  ${7.05} \mhide{\pm 0.073}$ & ${0.0106} \mhide{\pm 0.0005}$ \\
    Random Labels & 
    $240.0\mhide{\pm 12.02}$ & $2.90 \mhide{\pm 0.192}$ & ${29.04} \mhide{\pm 0.76}$ & ${6.97} \mhide{\pm 0.062}$ & ${0.0100} \mhide{\pm 0.0010}$ \\
    Ours + Online Clustering & 
    ${995.8} \mhide{\pm 0.86}$ & $0.17 \mhide{\pm 0.027}$ & ${31.56} \mhide{\pm 0.48}$ & ${6.82} \mhide{\pm 0.112}$ & ${0.0178} \mhide{\pm 0.0029}$ \\
    Ours & 
    $\textbf{1000.0} \mhide{\pm 0.00}$ & $0.08 \mhide{\pm 0.009}$ &  $\textbf{18.03} \pm 0.55$ & $\textbf{7.72} \pm 0.034$ & $\textbf{0.0015} \mhide{\pm 0.0004}$ \\
    \midrule
    Logo-GAN-RC~\cite{sage2018logo} & 
    $\textbf{1000.0} \mhide{\pm 0.00}$ & $0.08 \mhide{\pm 0.006}$ &  $28.83 \mhide{\pm 0.43}$ &  ${7.12} \mhide{\pm 0.047}$ & ${0.0091} \mhide{\pm 0.0001}$ \\
    Class-conditional GAN~\citep{mirza2014conditional} & 
    $\textbf{1000.0} \mhide{\pm 0.00}$ & $0.08 \mhide{\pm 0.003}$ &   ${23.56} \mhide{\pm 2.24}$ & ${7.44} \mhide{\pm 0.080}$ & ${0.0019} \mhide{\pm 0.0001}$ \\
    \bottomrule
  \end{tabular}%
} %
  \vspace{3pt}

  \vspace{-5pt}%
\end{table*}

\subsection{Synthetic Data Experiments}
\lblsec{toy}
The 2D-ring dataset is a mixture of 8 2D-Gaussians, with means $(\cos{\frac{2\pi i}{8}}, \sin{\frac{2\pi i}{8}})$ and variance $10^{-4}$, for $i \in \lbrace 0, \dots, 7 \rbrace$. 
The 2D-grid dataset is a mixture of 25 2D-Gaussians, each with means $(2i - 4, 2j- 4)$ and variance $0.0025$, for $i, j \in \lbrace 0, \dots, 4 \rbrace$. 

We follow the metrics used in prior work~\cite{guyon2017veegan,lin2018pacgan}. A generated point is deemed high-quality if it is within three standard deviations from some mean \citep{guyon2017veegan}. 
The number of modes covered by a generator is the number of means that have at least one corresponding high-quality point. 
To compare the reverse KL divergence between the generated distribution and the real distribution, we assign each point to its closest mean in Euclidean distance, and compare the empirical distributions \citep{lin2018pacgan}. 

Our results are displayed in Table \ref{2d_results}. 
We observe that our method generates both higher quality and more diverse samples than competing unsupervised methods. %
We also see that the success of our method is not solely due to the addition of class conditioning, as the conditional architecture with random labels still fails to capture diversity and quality. \new{Our online clustering variant, where we update the cluster means and memberships in an online fashion, is also able to perform well, achieving almost perfect performance. }

\subsection{Stacked-MNIST and CIFAR-10 Experiments}
\lblsec{mnist_cifar}
\begin{table}
    
 \centering%
\caption{\new{Fr\'{e}chet Inception Distance (FID) and Inception Score (IS) metrics for varying $k$ on CIFAR-10. Our method performs well for a wide range of $k$, but does worse for very small $k$ or very large $k$. Results are averaged over five trials, with standard error reported.}}
\label{k-dependence-cifar}
\vspace{-2mm}
\resizebox{%
  \ifdim\width>\columnwidth
    \columnwidth
  \else
    0.88\columnwidth
  \fi
}{!}
{%

\begin{tabular}{lrr}
\toprule
\multicolumn{1}{c}{} &
\multicolumn{2}{c}{CIFAR-10} \\
\cmidrule(r){2-3}
        & \multicolumn{1}{c}{FID $\downarrow$}  &  \multicolumn{1}{c}{IS $\uparrow$} \\
\midrule
GAN~\cite{goodfellow2014generative} & 
$28.08 \mhide{\pm 0.47}$ &  ${6.98} \mhide{\pm 0.062}$ \\
$k=10$  &  ${32.55} \pm 2.58$ & ${7.03} \pm 0.158$ \\
$k=25$  &  ${19.21} \pm 0.73$ & $\textbf{7.79} \pm 0.040$ \\
$k=100$ (default)  & $\textbf{18.03} \pm 0.55$ & ${7.72} \pm 0.034$\\
$k=250$  &  ${18.49} \pm 0.53$ & ${7.74} \pm 0.053$ \\
$k=1000$  &  ${20.76} \pm 0.26$ & ${7.48} \pm 0.028$ \\
\midrule
Class Conditional GAN~\citep{mirza2014conditional} & ${23.56} \mhide{\pm 2.24}$ & ${7.44} \mhide{\pm 0.080}$ \\
\bottomrule
\end{tabular}
} %
\vspace{-1pt}

\end{table}

The Stacked-MNIST dataset~\cite{lecun1998mnist,guyon2017veegan,lin2018pacgan} is produced by stacking three randomly sampled MNIST digits into an RGB image, one per channel, generating $10^3$ modes with high probability. 
To calculate our reverse KL metric, we use pre-trained MNIST and CIFAR-10 classifiers to classify and count the occurrences of each mode. %
For these experiments, we use $k=100$.
We also compare our method \new{with our} online clustering variant.

Our results are shown in Table~\ref{small_results}. On both datasets, we achieve large gains in diversity, significantly improving \fid (FID) and Inception Score (IS) over even supervised class-conditional GANs.

\new{We also conduct experiments to test how sensitive our method is to the choice of the number of clusters $k$. The results with varying $k$ values are shown in Table~\ref{k-dependence-cifar}. We observe that our method is robust to the choice of $k$, as long as it is not too low or too high. Choosing a $k$ too low reduces to unconditional GAN behavior, while choosing a $k$ too high potentially makes clustering unstable, hence hurting GAN training.}

\subsection{Large-Scale Image Datasets Experiments}
\lblsec{large-scale}
Lastly, we measure the quality and diversity of images for GANs trained on large-scale datasets \new{such as ImageNet and Places365}. We train using all 1.2 million ImageNet challenge images across all 1000 classes and all 1.8 million Places365 images across 365 classes.  No class labels were revealed to the model. For both datasets, we choose $k=100$.

In following sections, we present detailed analyses comparing our method with various baselines on realism, diversity, and mode dropping.
\subsubsection{Generation Realism and Diversity}

\begin{table*}%
    \vspace{-10pt}
  \centering%
   \caption{Fr\'{e}chet Inception Distance (FID), Fr\'{e}chet Segmentation Distance (FSD), and Inception Score (IS) metrics for Places365 and ImageNet experiments. Our method improves in both quality and diversity over previous methods but still fails to reach the quality of fully-supervised class conditional ones.} 
   \label{large_results}
    \vspace{-2.5mm}
    \resizebox{%
\ifdim\width>\textwidth
  \textwidth
\else
  0.88\width
\fi
}{!}
    {%
    
    \begin{tabular}{lrrrrrr}
    \toprule
    \multicolumn{1}{c}{} &
    \multicolumn{3}{c}{Places365} & 
    \multicolumn{3}{c}{ImageNet} \\
    \cmidrule(r){2-4}
    \cmidrule(r){5-7}
            & \multicolumn{1}{c}{FID $\downarrow$}  &  \multicolumn{1}{c}{FSD $\uparrow$}  & \multicolumn{1}{c}{IS $\uparrow$}  & \multicolumn{1}{c}{FID $\downarrow$}  &  \multicolumn{1}{c}{FSD $\uparrow$}  & \multicolumn{1}{c}{IS $\uparrow$}  \\
    \midrule
    GAN~\cite{goodfellow2014generative} & 14.21 & 125.4 & 8.71  & 54.17 & 129.7 & 14.01  \\
    PacGAN2~\cite{lin2018pacgan} & 18.02 & 161.1 & 8.58  & 57.51 & 171.9 & 13.50  \\
    PacGAN3~\cite{lin2018pacgan} & 22.00 & 221.7 & 8.56 & 66.97 & 190.5 & 12.34  \\
    \mhide{MGAN~\cite{hoang2018mgan}} & \mhide{15.78} & 156.2 & \mhide{8.41}  & \mhide{58.88} & 156.6 & \mhide{13.22}  \\
    \mhide{RotNet Feature Clustering} & \mhide{14.88} &131.4 & \mhide{8.54}  & \mhide{53.75} &126.8 & \mhide{13.76}  \\
    Logo-GAN-AE~\cite{sage2018logo}  & 14.51 & 140.2 & 8.19 & 50.90 & 119.7 & 14.44  \\
    Random Labels & 14.20 & 112.4 & 8.82  & 56.03 & 146.3 & 14.17  \\
    Ours  & \textbf{9.56} & \textbf{87.67} & \textbf{8.94}  & \textbf{40.30} & \textbf{82.46} & \textbf{15.82}  \\
    \midrule
    Logo-GAN-RC~\cite{sage2018logo} & 8.66 & 75.51 & 10.55  & 38.41 & 90.03 & 18.86  \\
    Class Conditional GAN~\citep{mirza2014conditional} & 8.12 & 58.17 & 10.97 & 35.14 & 115.1 & 24.04 \\
    \bottomrule
  \end{tabular}
  } %
  \vspace{-3pt}

\end{table*}

\paragraph{Realism metrics. }\new{Across datasets, our method improves the quality of generated images in terms of standard quality metrics such as FID and IS. We also evaluate the generated samples with the Fr\'{e}chet Segmentation Distance (FSD) \citep{bau2019seeing}, which measures discrepancies between the segmentation statistics of the generated samples and real samples. As shown in Table~\ref{large_results}, our method outperforms all the unsupervised baselines tested against on these metrics. Although Logo-GAN-RC outperforms our method, it is based on a supervised pre-trained ImageNet classifier. }

\myparagraph{Diversity visualizations. }\reffig{sample-diversity} shows samples of the generated output, comparing our model to a vanilla GAN and a PacGAN model and a sample of training images. 
To highlight differences in diversity between the different GAN outputs, for each compared method, we assign $50{,}000$ unconditionally generated images with labels given by standard ResNet50 \citep{he2016deep} classifiers trained on respective datasets \citep{zhou2014learning,torchvisionmodels}, and visualize images with highest classifier score for certain classes in \reffig{places-embassy}. 
We observe that across classes, vanilla GAN and PacGAN tend to synthesize less diverse samples, repeating many similar images. On the other hand, our method improves diversity significantly and produces images with varying color, lighting conditions, and backgrounds.

\subsubsection{Quantifying Mode Dropping with Image Reconstruction}
\begin{figure}[ht]%
    \centering%
    \includegraphics[width=0.93\columnwidth]{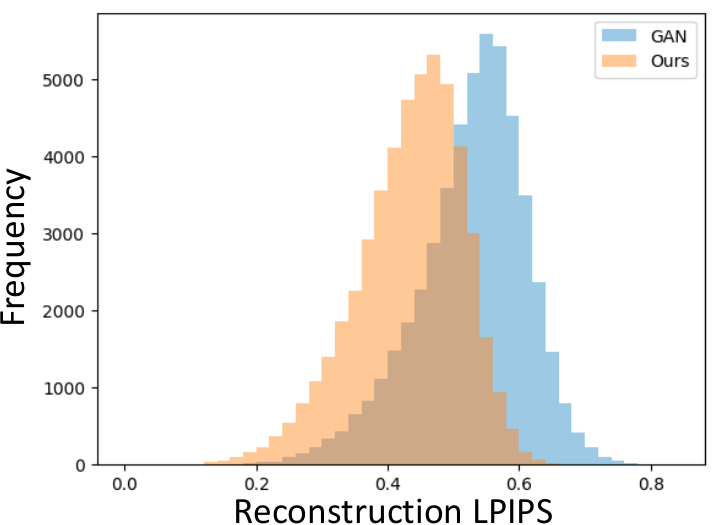}\vspace{-6pt}
    \caption{We calculate the reconstruction $\mathsf{LPIPS}$ error for $50{,}000$ randomly sampled training images. Overall, our method is able to achieve better reconstruction than a vanilla GAN. }
    \lblfig{cluster-lpips}
    \vspace{-3pt}
\end{figure}
\begin{figure*}[ht]
  \centering
  \begin{subfigure}[b]{\textwidth}
    \includegraphics[width=\textwidth]{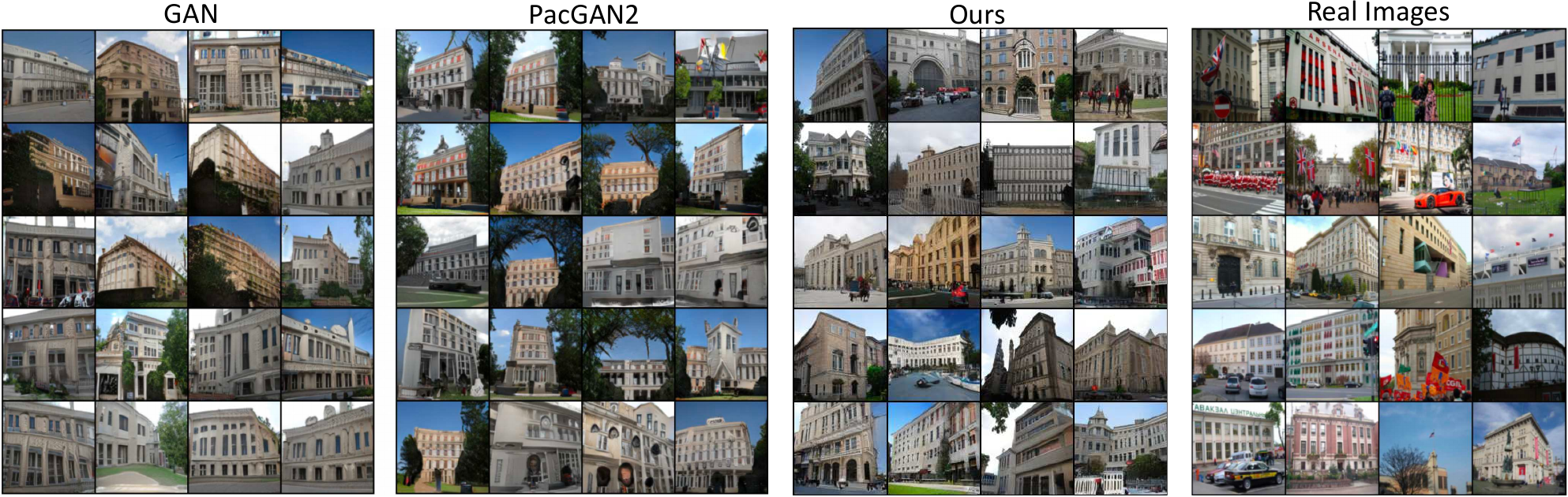}
    \caption{Places365 samples with high classifier scores on the ``\texttt{embassy}'' category.}\lblfig{places-embassy}
  \end{subfigure}\vspace{0.075cm}
  \begin{subfigure}[b]{\textwidth}
    \includegraphics[width=\textwidth]{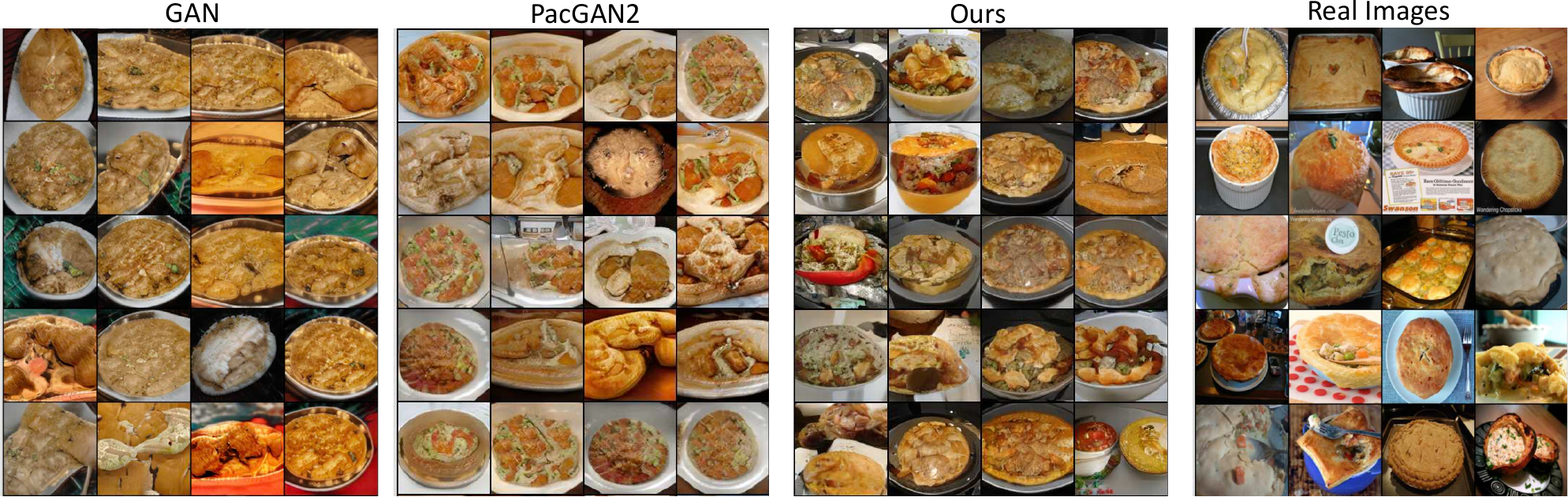}
    \caption{ImageNet samples with high classifier scores on the ``\texttt{pot pie}'' category.}\lblfig{imagenet-pot-pie}
  \end{subfigure}
  \vspace{-18pt}
  \caption{Comparing visual diversity between the samples from a vanilla unconditional GAN, PacGAN2, our method, and real images for specific categories. Images are sorted in decreasing order by classifier confidence on the class from top left to bottom right.  %
  Our method successfully increases sample diversity compared to vanilla GAN and PacGAN2.
  More visualizations are available in \refapp{app-additional-examples}.
  }
  \lblfig{sample-diversity}
  \vspace{-3pt}
\end{figure*}
\begin{figure*}
\centering
\vspace{-15pt}
\includegraphics[width=\textwidth]{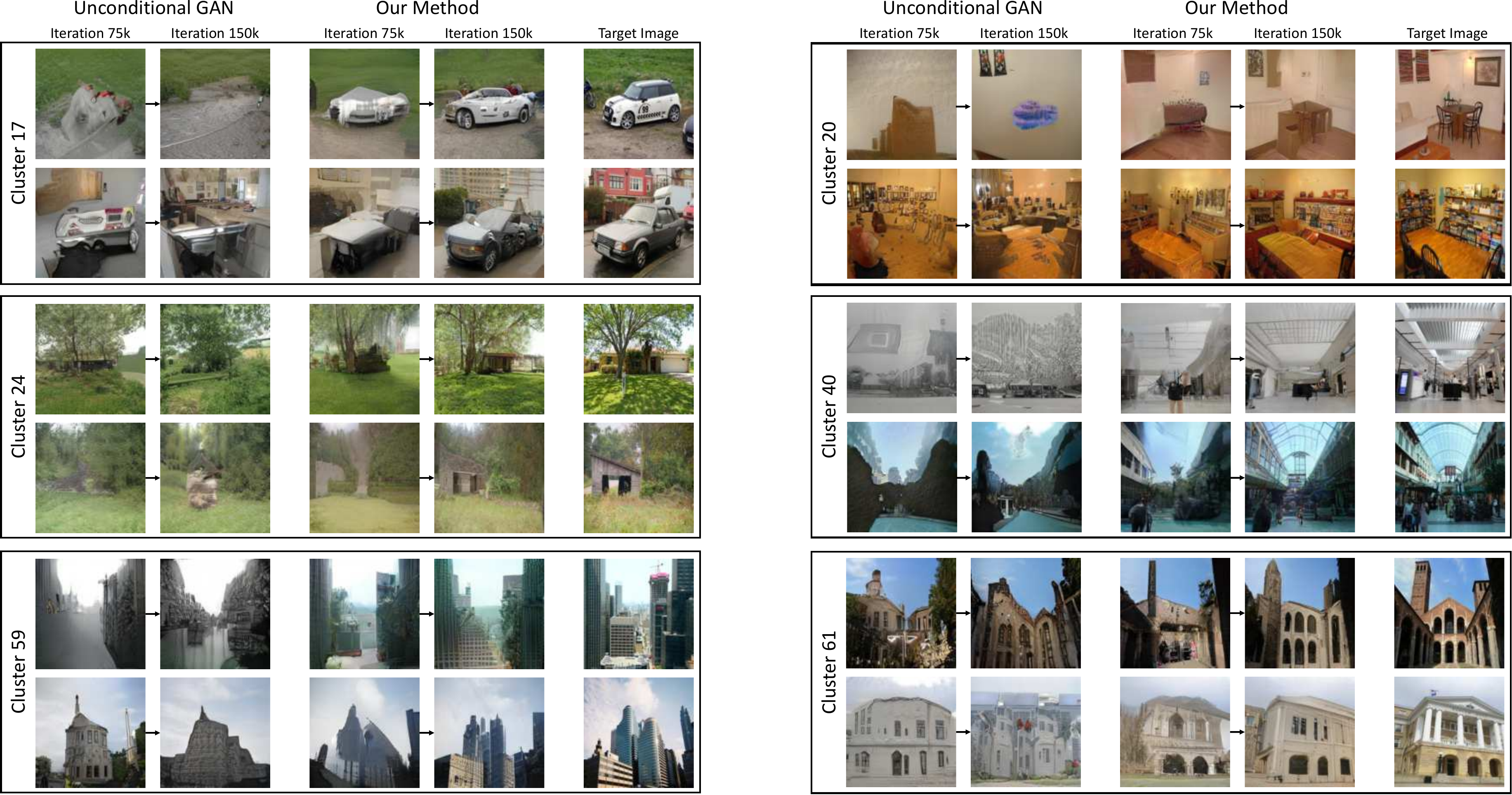}
\vspace{-17pt}
\caption{Improvements of GAN reconstructions during training.  Each GAN-generated image shown has been optimized to reconstruct a specific training set image from the Places365 dataset, at right.  Reconstructions by generators from an early training iteration of each model are compared with the final trained generators.  Self-conditioning the model results in improved synthesis of clustered features such as wheels, buildings, and indoor objects.
}
\lblfig{cluster-evolution}%
\vspace{-4pt}
\end{figure*}

While random samples reveal the capacity of the generator, reconstructions of training set images~\cite{zhu2016generative,brock2017neural} can be used to visualize the omissions (dropped modes) of the generator~\citep{bau2019seeing}.

Previous GAN inversion methods do not account for class conditioning, despite recent efforts by concurrent work~\cite{huh2020ganprojection}. Here we extend the encoder + optimization hybrid method~\cite{zhu2016generative,bau2019seeing}. We first train an encoder backbone $F\colon \rvx \rightarrow \rvr$ jointly with a classifier $F_c\colon \rvr \rightarrow c$ and a reconstruction network $F_z\colon \rvr \rightarrow \rvz$ to recover both the class $c$ and the original $\rvz$ of a generated image.  We then optimize $\rvz$ to match the pixels of the query image $\rvx$ as well as encoder features extracted by $F$:
\begin{equation}\label{eq:recperimg}
\mathcal{L_{\mathsf{rec}}}(\rvz, c) = \lVert G(\rvz, c) - \rvx\rVert_1  + \lambda_f  \lVert F(G(\rvz, c)) - F(\rvx)\rVert^2_2.
\end{equation}
We set $\lambda_f = 5\times10^{-4}$.  When initialized using $F_z$ and $F_c$, this optimization faithfully reconstructs images generated by $G$, and reconstruction errors of real images reveal cases that $G$ omits. More details regarding image reconstruction can be found in \refapp{app-reconstruction}.

To evaluate how well our model can reconstruct the data distribution, we compute the average $\mathsf{LPIPS}$ perceptual similarity score \citep{zhang2018perceptual} between $50{,}000$ ground truth images and their reconstructions. Between two images, a low $\mathsf{LPIPS}$ score suggests the reconstructed images are similar to target real images. We find that on Places365, our model is able to better reconstruct the real images, with an average $\mathsf{LPIPS}$ score of $0.433$, as compared to the baseline score of $0.528$. A distribution of the $\mathsf{LPIPS}$ reconstruction losses across the $50{,}000$ images can be visualized in \reffig{cluster-lpips}.

\reffig{cluster-evolution} visually shows how reconstruction quality evolves as we perform reclustering and training. %
Visualizations show our model improves reconstruction during training, and recovers distinctive features, including improved forms for cars, buildings, and indoor objects. On the other hand, the vanilla GAN struggles to reconstruct most of the real images throughout the training.

\subsection{Clustering Metrics}
\begin{table}%
  \centering%
  \caption{Normalized Mutual Information (NMI) and purity metrics for the clusters obtained by our method on various datasets.} 
\vspace{-2.5mm}
    \resizebox{%
      \ifdim\width>\columnwidth
        \columnwidth
      \else
        \width
      \fi
    }{!}
    {%
    \begin{tabular}{lrrrrrr}
    \toprule
            & \thead{2D \\ Ring}  &  \thead{2D \\ Grid } & \thead{Stacked \\ MNIST} & CIFAR-10 & Places365 & ImageNet \\
    \midrule
    NMI & 1.0 & 0.9716 & 0.3018 & 0.3326 & 0.1744 & 0.1739 \\
    Purity & 1.0 & 0.9921 & 0.1888 & 0.1173 & 0.1127 & 0.1293\\
    \bottomrule
  \end{tabular}
  } %
  \label{clustering_metrics}
\end{table}

We measure the quality of our clustering through Normalized Mutual Information (NMI) and clustering purity across all experiments in Table~\ref{clustering_metrics}. 

NMI is defined as $\mathsf{NMI}(X, Y) = \frac{2 I(X; Y)}{H(X) + H(Y)}$, where $I$ is mutual information and $H$ is entropy. 
NMI lies in $[0, 1]$, and higher NMI suggests higher quality of clustering. Purity is defined as $\frac{1}{N} \sum_c \max_y |\pi_c \cap \pi^*_y|$, where $\{\pi_c\}_{c=1}^k$ is the partition of inferred clusters and $\{\pi^*_y\}_{y=1}^k$ is the partition given by the true classes. Higher purity suggests higher clustering quality.
Purity is close to $1$ when each cluster has a large majority of points from some true class set.

We observe that many of our clusters in large-scale datasets do not correspond directly to true classes, but instead corresponded to object classes. 
For example, we see that many clusters corresponded to people and animals, none of which are part of a true class, which is an explanation for low clustering metric scores. 

Though our clustering scores are low, they are significantly better than a random clustering. Randomly clustering ImageNet to 100 clusters gives an NMI of $0.0069$ and a purity of $0.0176$; randomly clustering Places to 100 clusters gives an NMI of $0.0019$ and a purity of $0.0137$.

\subsection{\new{Ablation Studies on Cluster Initialization and Matching}}
\label{sec:ablations}
\begin{table}
    \centering%
   \caption{\new{Fr\'{e}chet Inception Distance (FID) and Inception Score (IS) metrics for CIFAR-10 ablations experiments. The full algorithm is not necessary when using R1 regularization, but necessary without regularization. Results are averaged over 5 trials, with standard error reported.}}
   \label{ablations_results}
   
  \vspace{-2.5mm}
   \resizebox{%
      \ifdim\width>\columnwidth
        \columnwidth
      \else
        \width
      \fi
    }{!}
    {%
    
    \begin{tabular}{lrrrr}
    \toprule
    \multicolumn{1}{c}{} &
    \multicolumn{2}{c}{Without R1} & 
    \multicolumn{2}{c}{With R1} \\ 
    \cmidrule(r){2-3}
    \cmidrule(r){4-5}
            & \multicolumn{1}{c}{FID $\downarrow$}  &  \multicolumn{1}{c}{IS $\uparrow$}  & \multicolumn{1}{c}{FID $\downarrow$}  &  \multicolumn{1}{c}{IS $\uparrow$} \\
    \midrule
    No Matching/Init & $25.38 \pm 1.41$ & $7.29 \pm 0.068$  & $22.70 \pm 0.48$ & $7.38 \pm 0.062$ \\
    No Matching & $18.91 \pm 0.65$ &  $7.51 \pm 0.073$  & $23.26 \pm 0.40$ & $7.27 \pm 0.029$ \\
    No Initialization & $19.85 \pm 0.65$ &  $7.58 \pm 0.046$  & $22.02 \pm 0.21$ & $7.35 \pm 0.024$\\
    Full Algorithm  & $\textbf{18.03} \pm 0.55$ & $\textbf{7.72} \pm 0.034$ & $22.80 \pm 0.27$ & $7.32 \pm 0.031$  \\
    \bottomrule
  \end{tabular}
  } %

\end{table}

\new{
We perform ablation studies on the cluster initialization and cluster matching used in our full algorithm on CIFAR-10. The results are summarized in Table~\ref{ablations_results}}.

\new{
We find that these clustering tricks are required for stable clustering and ultimately for GAN quality. The less stable the clustering is, the worse the GAN performs. To quantify cluster stability, we measure the NMI between a given clustering and its previous clustering. We observe that without the two clustering tricks, the clustering is not stable, achieving an NMI of $0.55$ compared to the full algorithm's $0.74$. }

\new{In contrast, with R1 regularization \citep{mescheder2018training}, our method still works well on CIFAR-10 without cluster initialization and matching. We suspect that this is because the GAN regularization loss stabilizes GAN training, preventing cluster instability from hurting GAN training stability. However, our results on CIFAR-10 worsen when adding R1 regularization. We hypothesize that this is due to discriminator regularization decreasing the quality of the discriminator's features, which ultimately leads to worse clusters. We investigate this interaction further in \refapp{app-reg}.}

\section{Conclusion}
We have found that when a conditional GAN is trained with clustering labels derived from discriminator features, it is effective at reducing mode collapse, outperforming several previous approaches. We observe that the method continues to perform well when the number of synthesized labels exceeds the number of modes in the data. Furthermore, our method scales well to large-scale datasets, improving various standard measures on ImageNet and Places365 generation, and generating images that are qualitatively more diverse than many unconditional GAN methods.

\myparagraph{Acknowledgments.}
We thank Phillip Isola, Bryan Russell, Richard Zhang, and our anonymous reviewers for helpful comments. Tongzhou Wang was supported by the MIT EECS Merrill Lynch Graduate Fellowship. We are grateful for the support from the DARPA XAI program FA8750-18-C0004, NSF 1524817, NSF BIGDATA 1447476, and a GPU donation from NVIDIA.

\clearpage
{\small
\bibliographystyle{ieee_fullname}
\bibliography{reference}
}

\clearpage

\begin{appendices}

\section{Additional Experiment details}
\lblsec{app-details}
\myparagraphfirst{Compute resource details. }
For all experiments except ImageNet \citep{deng2009imagenet} and Places365 \citep{zhou2017scene} experiments, we used a single Titan Xp GPU. For ImageNet and Places365 experiments, we used four Titan Xp GPUs.

\myparagraph{Classifier details.}
We use a classifier to compute the reverse-KL metric for experiments on synthetic data, Stacked-MNIST, and CIFAR. 
To classify points for 2D experiments, we classify each point as the label of its nearest mode. 
To classify samples for Stacked-MNIST experiments, we use the pretrained MNIST classifier on each channel of the generated image. This gives us $3$ labels in $0, \dots, 9$, which is used to assign the image a label from $0, \dots, 999$. 
To classify samples for CIFAR-10, we a pretrained CIFAR-10 classifier. Both the MNIST and CIFAR-10 classifiers are taken from a pretrained model 
\href{https://github.com/aaron-xichen/pytorch-playground}{repository}. 

We also use a classifier to visualize the diversity of the trained models on ImageNet and Places. For ImageNet, we use a pretrained ResNet-50 obtained through the \texttt{torchvision} \citep{torchvisionmodels} package. For Places, we use a pretrained ResNet-50 released by \citep{zhou2017places}. 

\myparagraph{Clustering details.}
To compute cluster centers, for Stacked-MNIST and CIFAR-10 experiments, we cluster a random subset of $25{,}000$ images from the training set, and for ImageNet and Places365 experiments, we cluster a random subset of $50{,}000$ images from the training set. On synthetic datasets, we cluster a random sample of $10{,}000$ data points. While conducting $k$-means++, we cluster the subset ten times and choose the clustering that obtains the best performance on the clustering objective. We use the outputs of the last hidden layer of the discriminator as features for clustering. 

\myparagraph{Realism metrics Details.}
 All FIDs, FSDs, and Inception Scores (IS) are reported using $50{,}000$ samples from the fully trained models, with the FID and FSD computed using the training set. No truncation trick is used to sample from the generator. %
\myparagraph{Reverse KL Details. }
In all reverse-KL numbers, we report $D_{KL}(p_{\mathsf{fake}}\hspace{2pt}||\hspace{2pt}p_{\mathsf{real}})$, and treat $0 \log 0 = 0$, where we obtain the distributions for a set of samples by using the empirical distribution of classifier predictions.

\myparagraph{Online clustering details.}
As a hyperparameter study, we implement an online variant of our algorithm, where we update our clusters every training iteration. We train a model with a fixed clustering for $25{,}000$ iterations, then apply the online $k$-means algorithm of \citep{bottou1995convergence} using the training mini-batches as the $k$-means mini-batches. We found that while the online variant was able to do well on easier datasets such as 2D-grid and Stacked-MNIST, they did not achieve convincing performance on real images. 

\myparagraph{Logo-GAN details.}
For Logo-GAN experiments, we infer labels by clustering the outputs of the last layer of a pretrained ResNet50 ImageNet classifier. We train another model where labels are from clustering an autoencoder's latent space. The autoencoder architecture is identical to an unconditional GAN architecture, except that the latent space of the autoencoder is $512$. The autoencoders were trained to optimize a mean squared loss using Adam with a learning rate of $10^{-4}$ with a batch size of $128$, for $200{,}000$ iterations. For both these methods, we cluster a batch of $50{,}000$ outputs into $k=100$ clusters using the $k$-means++ algorithm.

\myparagraph{MGAN details.}
For MGAN experiments with synthetic Gaussian data, we use $\beta = 0.125$, following the original paper \citep{hoang2018mgan} on similar typed data.

\section{Synthetic Data Experiments}
\subsection{Comparison with MGAN}
\myparagraphfirst{Choice of $k$.}
We compare our method's sensitivity regarding the number of clusters $k$ to existing clustering-based mode collapse methods \citep{hoang2018mgan}.  

\begin{figure*}%
    \centering%
    \includegraphics[width=0.95\textwidth]{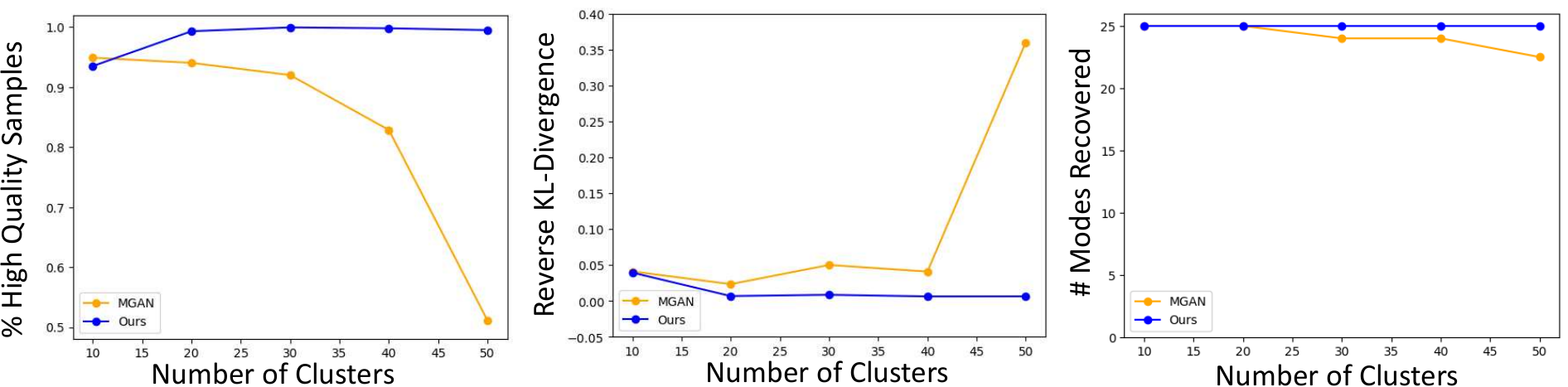}\vspace{-7pt}
    \caption{The dependence of our method and MGAN on the choice of number of clusters $k$ for 2D-grid (mixture of 25 Gaussians) experiment. MGAN is sensitive to the value $k$ and performance degrades if $k$ is too large. Our method is generally more stable and scales well with $k$.}
    \label{k_dependence}
\end{figure*}

For $k$ chosen to be larger than the ground truth number of modes, we observe that our method levels off and generates both diverse and high-quality samples.  
For $k$ chosen to be smaller than the ground truth number of modes, our results worsen. Figure \ref{k_dependence} plots the sensitivity of $k$ on the 2D-grid dataset, and the same trend holds for the 2D-ring dataset. 
Therefore, we lean towards using a larger number of clusters to ensure coverage of all modes.

MGAN~\citep{hoang2018mgan} learns a mixture of $k$ generators that can be compared to our $k$-way conditional generator. In Figure~\ref{k_dependence}, we see that MGAN performs worse as $k$ increases. We hypothesize that when $k$ is large, multiple generators must contribute to a single mode. Therefore, MGAN's auxiliary classification loss, which encourages each generator's output to be distinguishable, makes it harder for the generators to cover a mode collaboratively. On the other hand, our method scales well with $k$, because it dynamically updates cluster weights, and does not explicitly require the conditioned generators to output distinct distributions.

We run both our method and MGAN with varying $k$ values on the 2D-grid dataset. The results summarized in Figure~\ref{k_dependence} confirm our hypothesis that MGAN is sensitive to the choice of $k$, while our method is more stable and scales well with $k$. 

\myparagraph{Variance of data.} 
\begin{figure*}%
    \vspace{-1mm}
    \centering%
    \includegraphics[width=\textwidth]{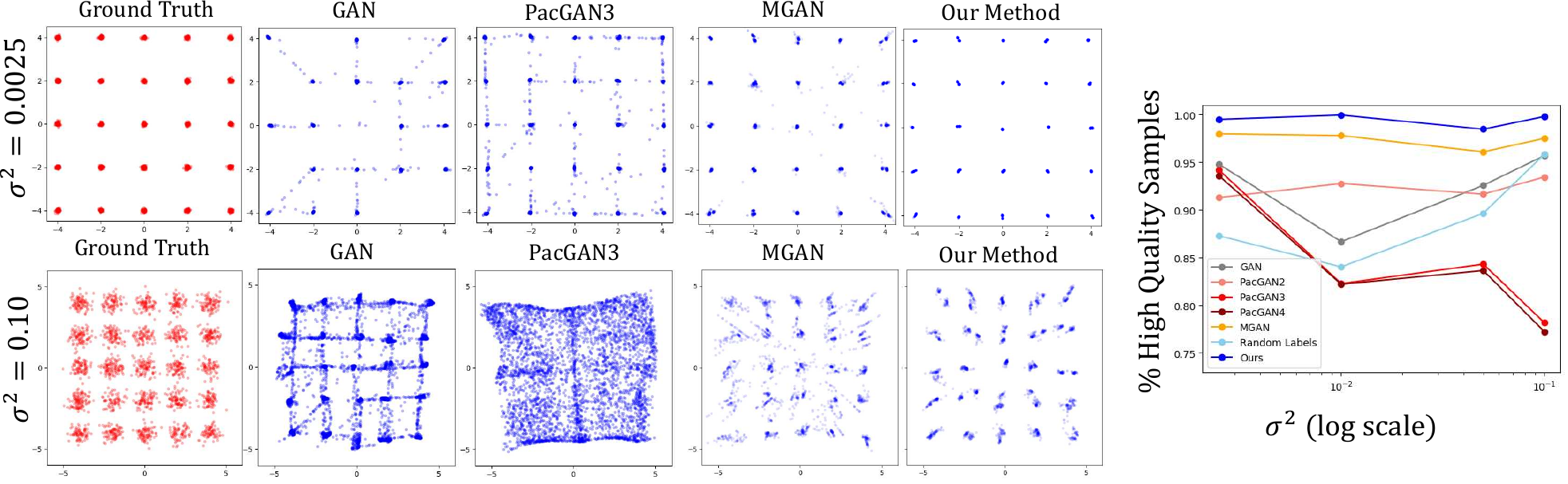}\vspace{-5pt}
    \caption{Visual comparison of generated samples on the 2D-grid dataset. Our method successfully covers all modes and generates high quality samples for low variance Gaussians. For high variance Gaussians, PacGAN learns a uniform distribution, while our method generates reasonably high quality samples. In these experiments, we used $k = 30$ for both our methods and MGAN.}
    \label{variance_test}
    \vspace{-5pt}
\end{figure*}

We compare our method and MGAN on the 2D-grid mixture of Gaussians on the sensitivity to the variance of each Gaussian. In Figure~\ref{variance_test}, when given $k=30$ to cover a mixture of $25$ Gaussians, MGAN does well in the large variance $\sigma^2 = 0.1$ experiment, but misses a mode in the small variance $\sigma^2 = 0.0025$ one, as the generators have difficulty working together on a single mode.

\section{Large-Scale Image Datasets}
\subsection{Hyperparameter Details}
\lblsec{app-large-details}

For experiments on synthetic datasets, we use $2$-dimensional latents, and train for $400$ epochs using Adam~\citep{kingma2014adam} with a batch size $100$ and learning rate $10^{-3}$. The embedding layer used for conditioning the generator has an output dimension of $32$. 

For CIFAR-10 and Stacked-MNIST experiments, we use $128$ latent dimensions, and Adam with a batch size of $64$ and a learning rate of $10^{-4}$ with $\beta_1=0.0, \beta_2=0.99$. We train for $50$ epochs on Stacked-MNIST and $400$ epochs on CIFAR-10. The generator embedding is  $256$-dimensional. 

For Places365 and ImageNet experiments, we train our networks from scratch using Adam with a batch size of 128, learning rate of $10^{-4}$, \new{$\beta_1 = 0.0$, and $\beta_2 = 0.99$} for 200,000 iterations. The generator embedding is $256$-dimensional.

\subsection{Combined ImageNet and Places365 Experiments}
\lblsec{app-combined}

\begin{table}%

  \centering%
   \caption{\new{Fr\'{e}chet Inception Distance (FID), Fr\'{e}chet Segmentation Distance (FSD), and Inception Score (IS) metrics for GANs trained on the combined datasets of Places365 and ImageNet. Our method improves in both quality and diversity over both vanilla unconditional GANs and class-conditional ones.}}
   \label{both_datasets_results}
  \vspace{-2.5mm}
    \resizebox{%
      \ifdim\width>\columnwidth
        \columnwidth
      \else
        0.88\columnwidth
      \fi
    }{!}
    {%

    \begin{tabular}{lrrr}
    \toprule
    \multicolumn{1}{c}{} &
    \multicolumn{3}{c}{Places365 + ImageNet} \\
    \cmidrule(r){2-4}
            & \multicolumn{1}{c}{FID $\downarrow$}  &  \multicolumn{1}{c}{FSD $\downarrow$}  & \multicolumn{1}{c}{IS $\uparrow$} \\
    \midrule
    GAN~\cite{goodfellow2014generative} & 36.78 & 167.1 & 11.0384    \\
    Ours  & \textbf{20.88} & \textbf{68.55} & \textbf{12.0632} \\
    \midrule
    Class Conditional GAN~\citep{mirza2014conditional} & 22.05 & 92.43 & {14.7734}  \\
    \bottomrule
  \end{tabular}
  } %

\end{table}
\new{We also test our method on the combined ImageNet and Places365 datasets and compare the results to those attained by an unconditional GAN and a class-conditional one. We still use $k=100$ for our method, but change the regularization on the unconditional GAN to $\gamma = 100$ for convergence. The results are summarized in Table~\ref{both_datasets_results}. Surprisingly, our method is able to outperform class-conditioned GANs in terms of FID and FSD.}

For these experiments, we construct the dataset by treating the classes of ImageNet and Places365 as separate. The combined dataset has roughly 3 million images with $1365$ total classes. It is important to note that the classes in this dataset are heavily imbalanced, with the 365 classes from Places365 each containing more images than the 1000 ImageNet classes. Thus, to sample from the class-conditional GAN, we do not sample labels from a uniform distribution, but instead from the distribution of class labels appropriately weighted towards the Places365 classes.

\subsection{Reconstruction of Real Images} 
\lblsec{app-reconstruction}

To train the encoder backbone for the self-conditioned GAN, we follow an analogous procedure to the GAN-seeing work \citep{bau2019seeing}, using a similar encoder architecture to GAN-stability's \citep{mescheder2018training} ResNet2 discriminator architecture. We invert the last seven layers separately in a layerwise fashion, then use these layerwise encoders jointly as initialization for the overall encoder. 

Given a fake image $x$ generated from latent $z$ and class $c$ with $G = G_f \circ G_i$ where $G_i \colon z, c \to r$, the encoder $E = E_f \circ E_i$ outputs a latent vector $\hat{z}$ and a class embedding $\hat{e_c}$, where $E_i \colon x \to r$ and is optimized for the loss function
\begin{multline}
    \mathcal{L}_{\mathsf{enc}} = \lVert \hat{z} - z \rVert_2 + \lVert G(\hat{z}, c) - x\rVert_2 \\ + ||G_i(z, c) - E_i(x)||_2 + \mathcal{L}_{\mathsf{classification}}(\hat{e_c}, c)
\end{multline}

This loss function is identical to the one used in the GAN-seeing work, except with an additional classification loss and an altered reconstruction loss in pixel space. 

The classification loss, $\mathcal{L}_{\mathsf{classification}}(\hat{e_c}, c)$, arises from the encoder additionally predicting the class embedding vector of a generated image. To obtain logits from an embedding prediction, we multiply it with the normalized true embedding vectors of the generator. The classification loss is the standard cross-entropy loss between these logits and the true cluster index. 

To measure reconstruction loss in pixel space, we take the ground truth embedding vector and predicted latent vector as input to the generator, and measure the mean squared error between the reconstructed image and the original image. We found that using the ground truth embedding vector, as opposed to the predicted embedding vector, improves inversion quality. 

To optimize the latent for a specific image, we follow Equation~\ref{eq:recperimg}, where the class is inferred by taking the index of the largest logit in the model's latent prediction. We use the output of the seventh layer of the encoder as features for the feature space loss.

\subsection{Visualizing Generators with GAN Dissection}
\lblsec{app-dissection}

  \begin{figure}[t]
  \centering
    \includegraphics[width=0.95\columnwidth]{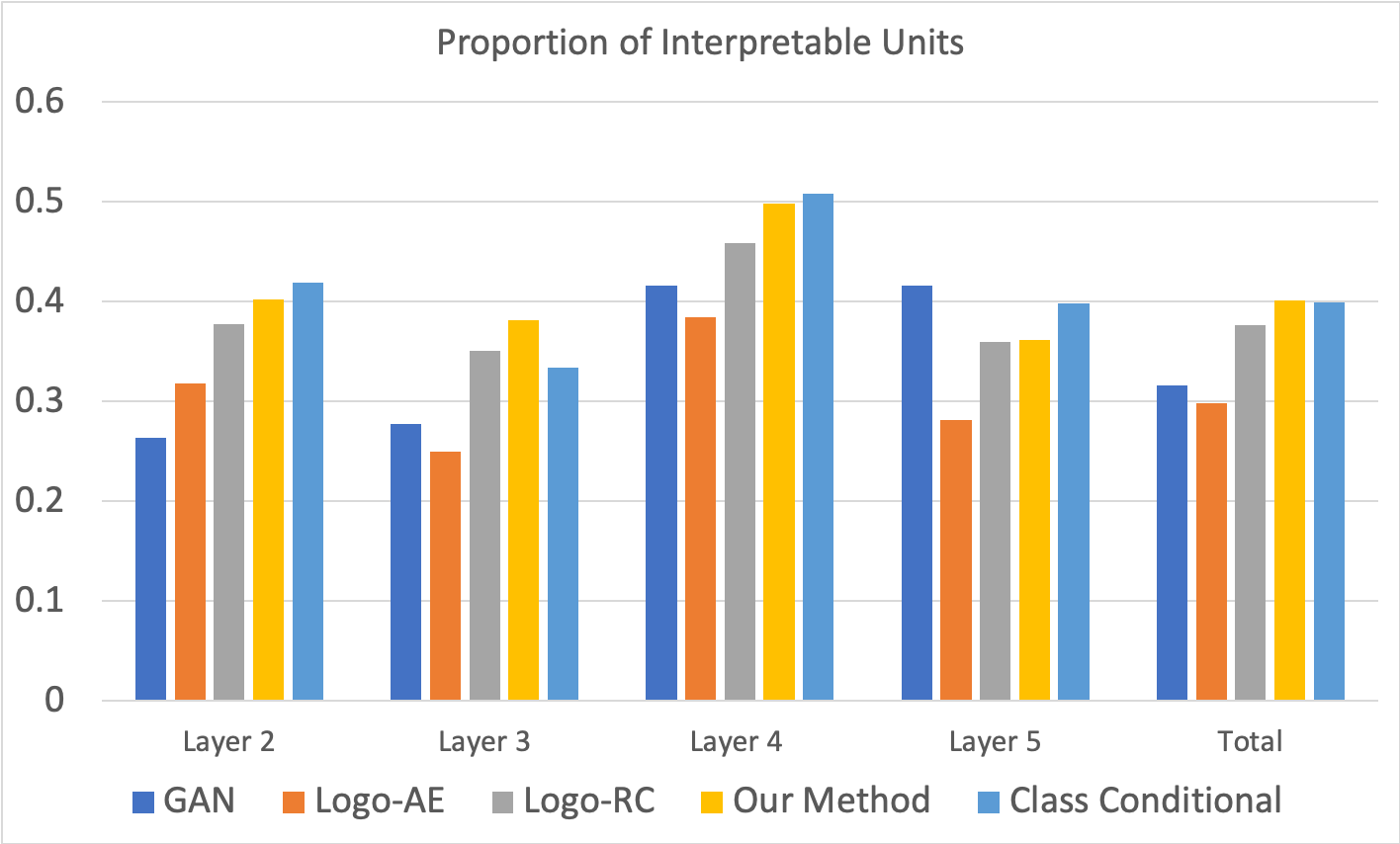}
   \caption{GAN Dissection results. We measure the proportion of interpretable units of generators trained on Places. We dissect the early to mid layers of the generator and count the number of units with IoU greater than 0.04.} 
  \lblfig{num-interpretable}
\end{figure}

\new{
We study the number of interpretable units in each learned generator. Previous work \citep{bau2019gandissect} has shown that a high number of interpretable units is positively correlated with the performance of GANs. Using the GAN Dissection method~\citep{bau2019gandissect}, we analyze each generator and count the number of units in each layer with IoU scores greater than $0.04$. As shown in \reffig{num-interpretable}, our generator has more interpretable units, on par with class-conditional models. 
}

\new{
We also visualize the internal mechanism of a general class-conditioned GAN. Using the GAN Dissection method~\citep{bau2019gandissect}, we dissect our generator conditioned on a specific cluster. With this, we are able to visualize each unit's concept when conditioned on a specific cluster. By monitoring the interpretable units, we can observe how the behavior of each unit changes as we change the conditioned cluster. We again deem a unit interpretable if it has an IoU score of higher than $0.04$. 
}

\new{
We observe that surprisingly, some units in the generator are responsible for different concepts when conditioning on different clusters. For example, the same units which are responsible for drawing people in one cluster are also responsible for drawing trees in a different cluster.}
\new{
There is also a significant amount of parameter sharing: units are reused to encode the same concept across different conditions. For example, water units and sky units tend to be reused across several conditions.}
\new{
The same phenomena in supervised class-conditional GANs. However, across the 20 conditions tested, we find that the self-conditioned GAN has a higher proportion of units whose concepts change across conditions. For a layer consisting of $1024$ units, we found $663$ self-conditioned GAN units change meaning across conditions, compared to $562$ units changing for a class-conditioned GAN. We can visualize these phenomena for both GANs on Places365 in \reffig{dissect}. 
}
\begin{figure*}
\centering
\caption{\new{Some units of the self-conditioned GAN (top) and a class-conditioned GAN (bottom) trained on Places365 correspond to different concepts when conditioned on different clusters (left), while other units correspond to the same concept across conditions (right). }}

\includegraphics[width=\textwidth]{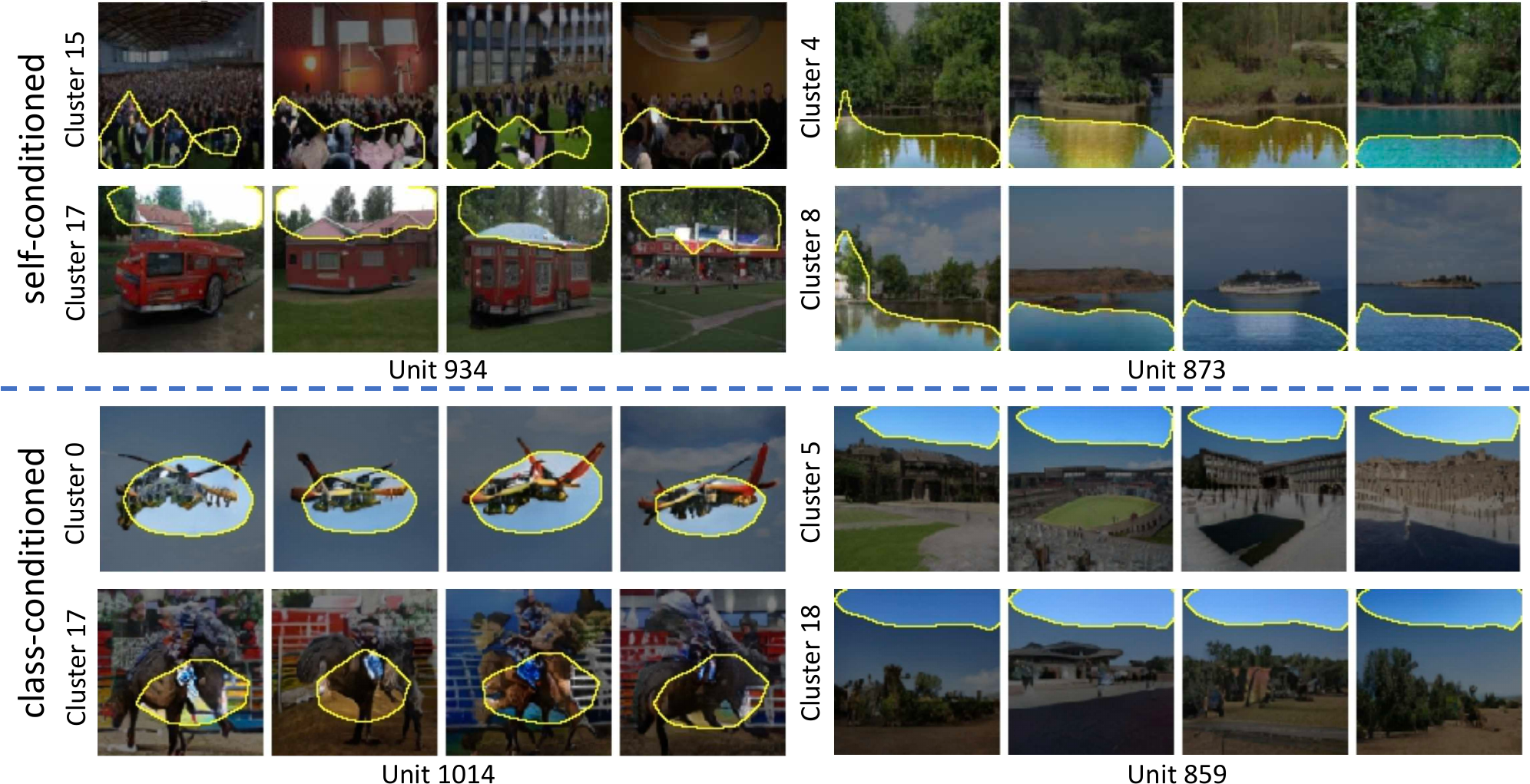}

\lblfig{dissect}%
\end{figure*}

\subsection{Interaction with discriminator regularization}
\lblsec{app-reg}
We find that our method's performance suffers if regularization is too strong and that our method performs best with weak regularization. To test this, we turn off the regularization in the experiments of \refsec{large-scale} and train our method, as well as unconditional and class-conditional baselines, using a smaller architecture (half number of layers compared to the ResNet-based generator we adapt from \citep{mescheder2018training}). Our method is able to match the performance of supervised models on ImageNet and Places365, achieving FID 22.68 on Places365 and FID 55.26 on ImageNet, compared to the class-conditional scores of 29.33 on Places365 and 53.36 on ImageNet and the unconditional scores of 75.09 on Places365 and 87.24 on ImageNet. 

We hypothesize that this is due to discriminator regularization leading to worse discriminator features to cluster with. To measure this, we train a linear ImageNet classifier over the last layer features of an R1-regularized ImageNet discriminator and similarly so for an unregularized ImageNet discriminator. Both discriminators in this case are fully class-conditional, meaning that they have seen true class labels throughout the training. We find that the classifier trained on top of the R1-regularized discriminator features severely underperforms the classifier trained on the unregularized discriminator features, achieving $30.7\%$ top-1 accuracy compared to $59.8\%$. The features chosen were the penultimate layer's outputs, and the linear classifier was trained using a standard cross-entropy loss with the Adam optimizer. We use a learning rate $10^{-4}$ with a batch size of $128$, $\beta_1 = 0.9$, and $\beta_2 = 0.99$.

\section{Additional Qualitative Results} 
\lblsec{app-additional-examples}
\begin{figure*}[ht]

\centering
\vspace{-17pt}
\begin{subfigure}[b]{\textwidth}
\includegraphics[width=\textwidth]{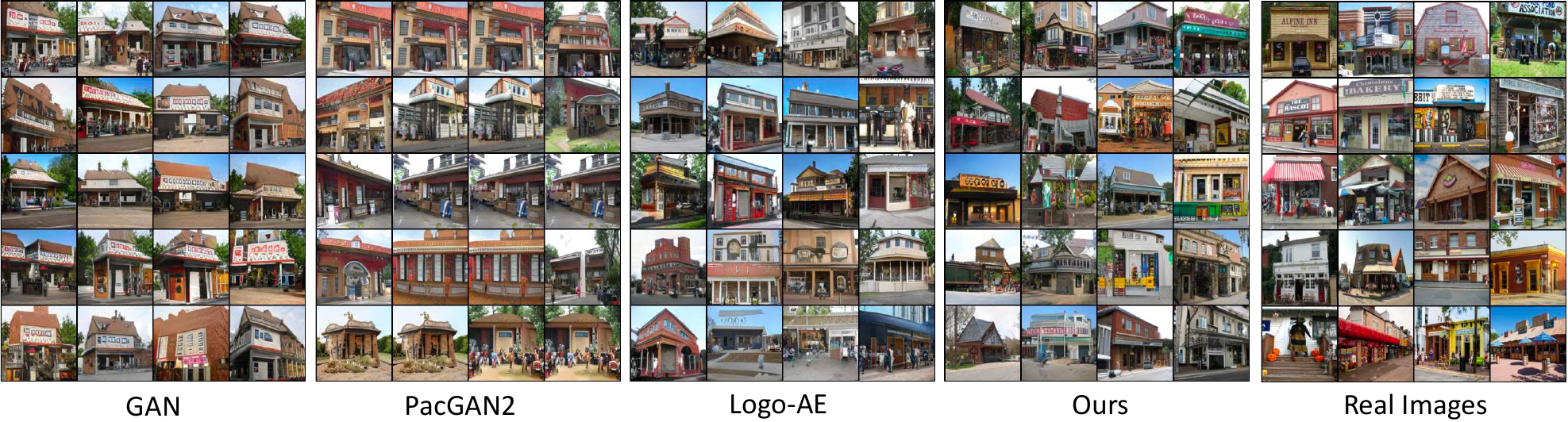}
\vspace{-17pt}
\caption{Places365 samples with high classifier scores on the ``\texttt{general store outdoor}'' category.}
\end{subfigure}
\begin{subfigure}[b]{\textwidth}
\includegraphics[width=\textwidth]{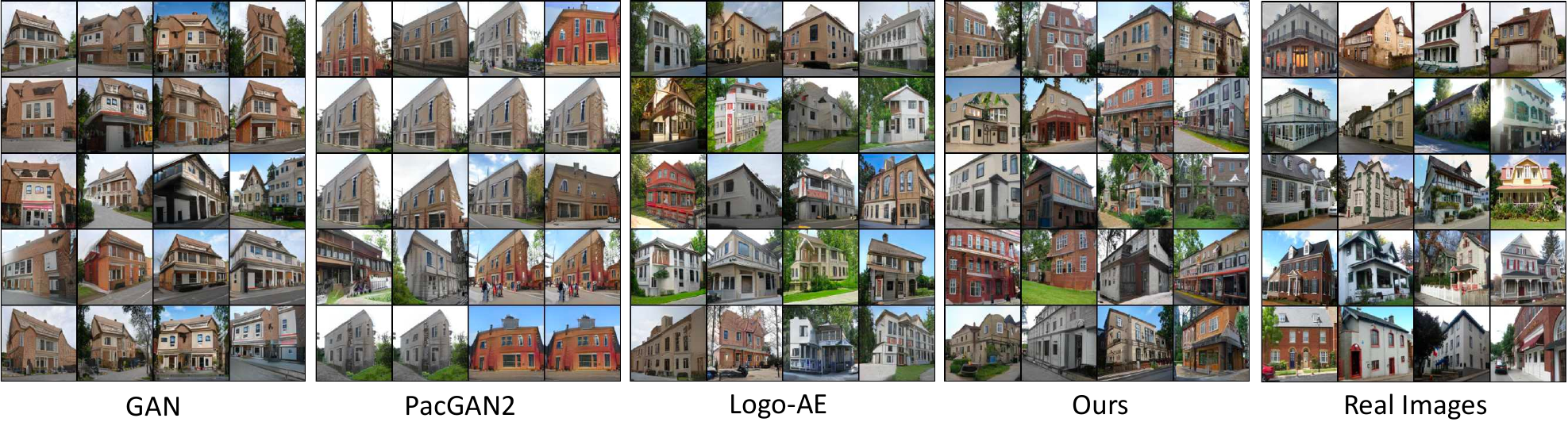}
\vspace{-17pt}
\caption{Places365 samples with high classifier scores on the ``\texttt{inn outdoor}'' category.}
\end{subfigure}
\begin{subfigure}[b]{\textwidth}
\includegraphics[width=\textwidth]{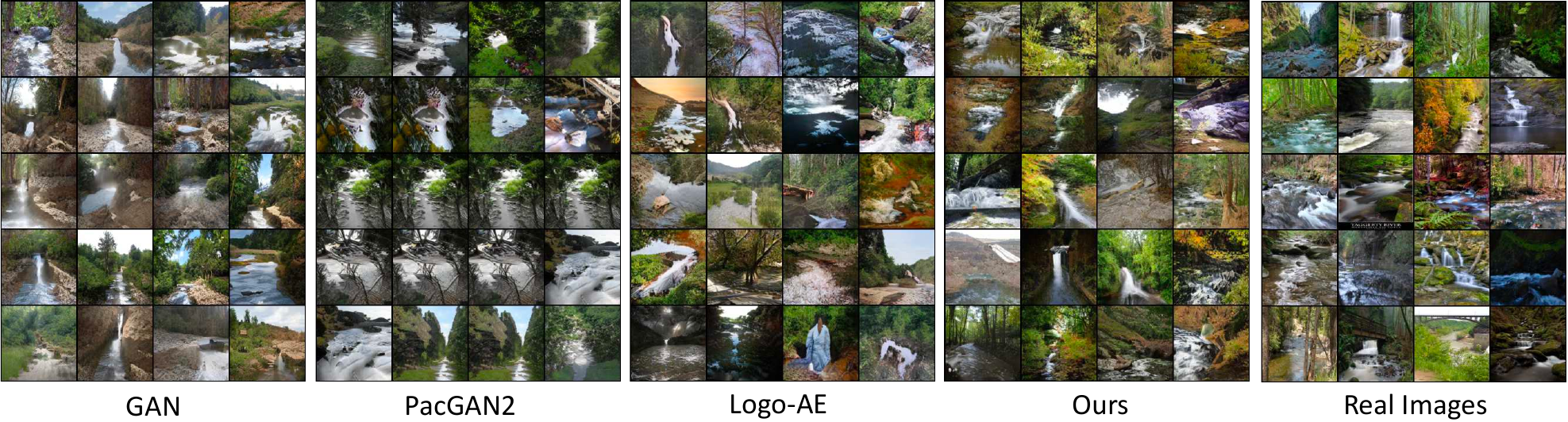}
\vspace{-17pt}
\caption{Places365 samples with high classifier scores on the ``\texttt{river}'' category.}
\end{subfigure}
\begin{subfigure}[b]{\textwidth}
\includegraphics[width=\textwidth]{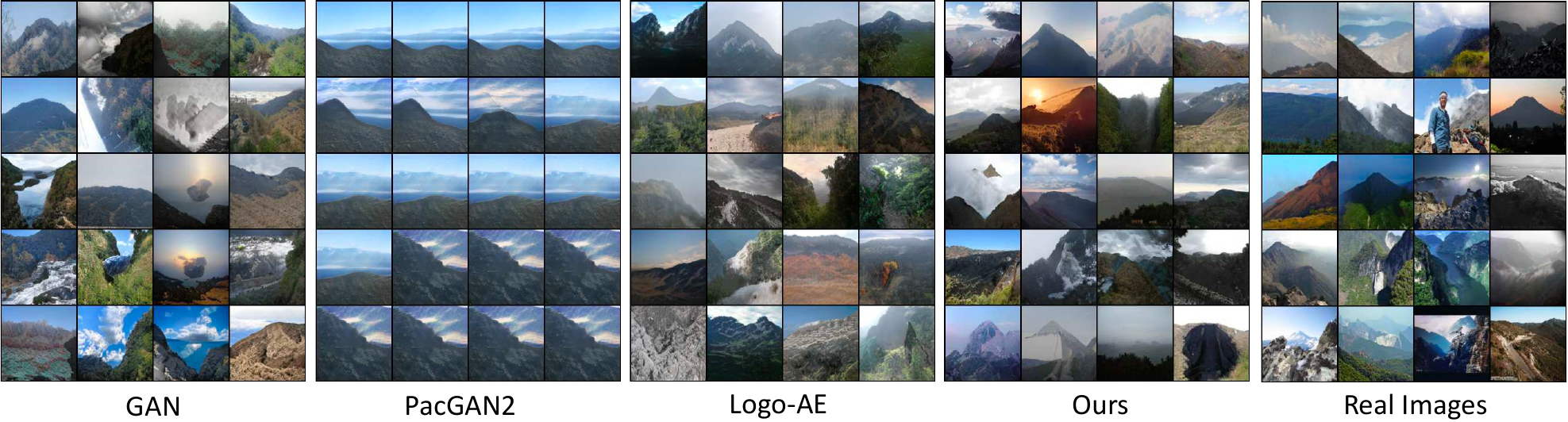}
\vspace{-17pt}
\caption{Places365 samples with high classifier scores on the ``\texttt{mountain}'' category.}
\end{subfigure}
\vspace{-19pt}
\caption{Places365~\cite{zhou2017scene} samples from an unconditional GAN, PacGAN2~\cite{lin2018pacgan}, Logo-GAN AE~\cite{sage2018logo}, our self-conditioned GAN, and real images.  We observe that competing methods exhibit severe mode collapse, frequently producing similar images with a few patterns, whereas our method captures a wider range of diversity.  Samples are all sorted in rank order of a classifier score for a single category.  Please refer to our \href{http://selfcondgan.csail.mit.edu/}{website} for additional examples. }%
\lblfig{supp-diversity-places}
\end{figure*}

\begin{figure*}[ht]\centering

\vspace{-17pt}
\begin{subfigure}[b]{\textwidth}
\includegraphics[width=\textwidth]{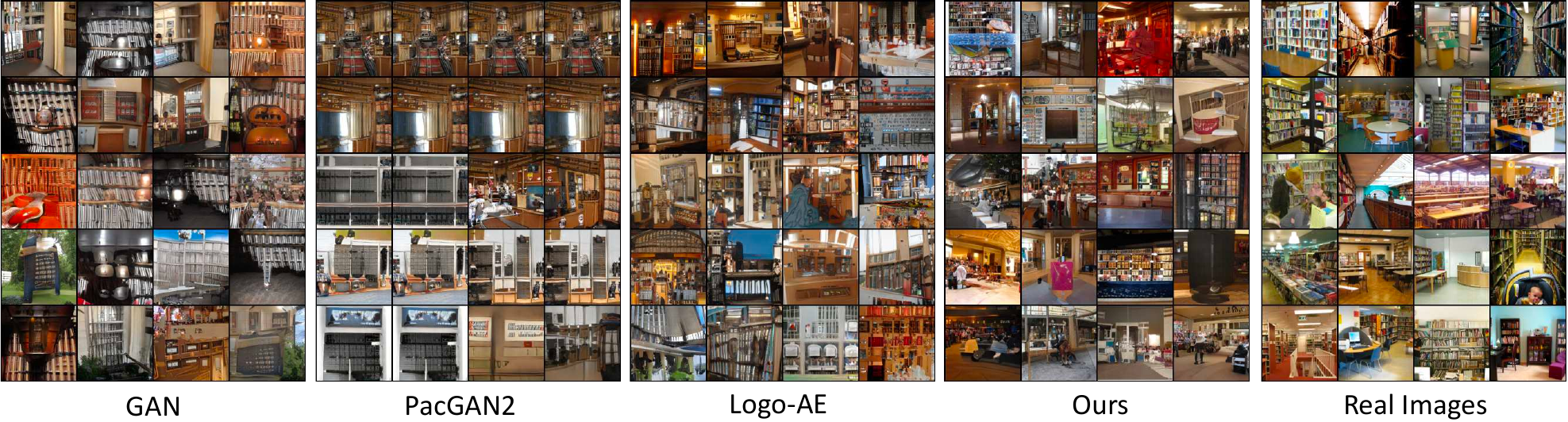}
\vspace{-17pt}
\caption{ImageNet samples with high classifier scores on the ``\texttt{library}'' category.}
\end{subfigure}
\begin{subfigure}[b]{\textwidth}
\includegraphics[width=\textwidth]{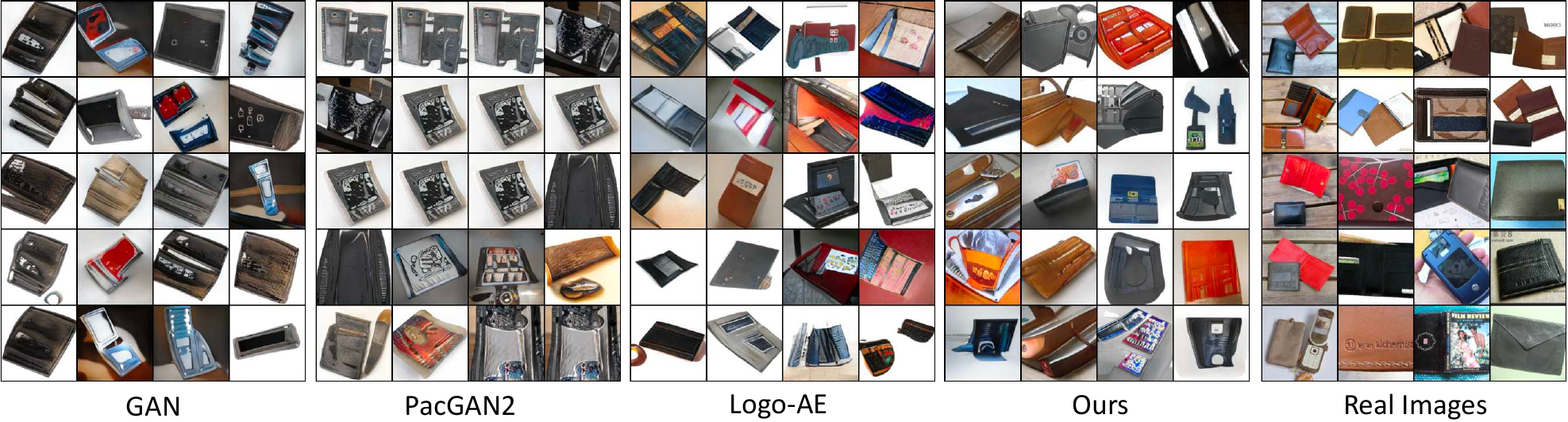}
\vspace{-17pt}
\caption{ImageNet samples with high classifier scores on the ``\texttt{wallet}'' category.}
\end{subfigure}
\begin{subfigure}[b]{\textwidth}
\includegraphics[width=\textwidth]{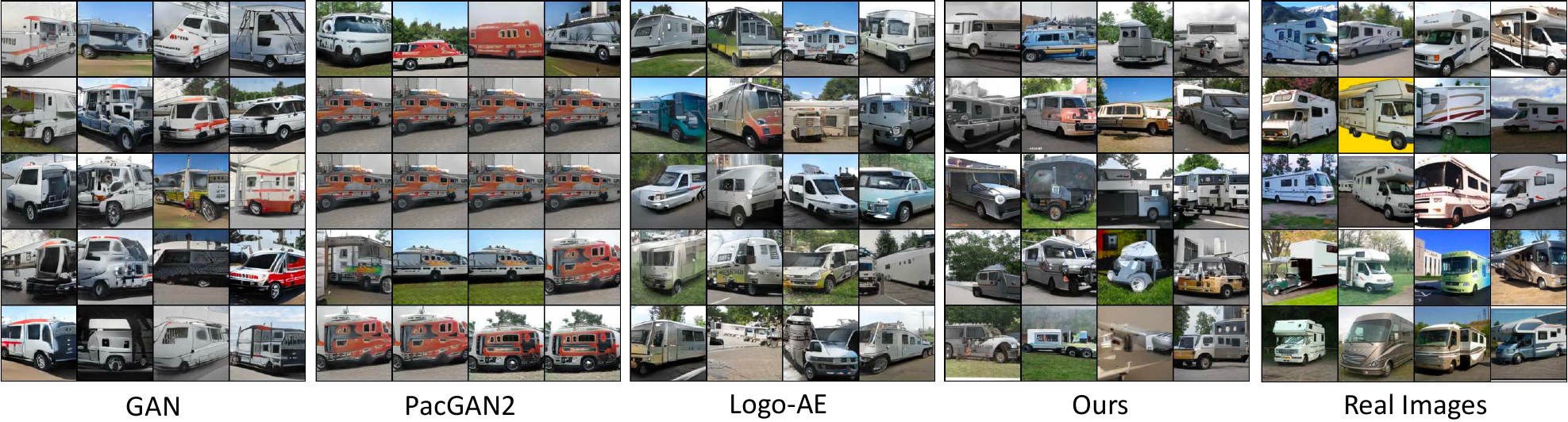}
\vspace{-17pt}
\caption{ImageNet samples with high classifier scores on the ``\texttt{recreational vehicle}'' category.}
\end{subfigure}
\begin{subfigure}[b]{\textwidth}
\includegraphics[width=\textwidth]{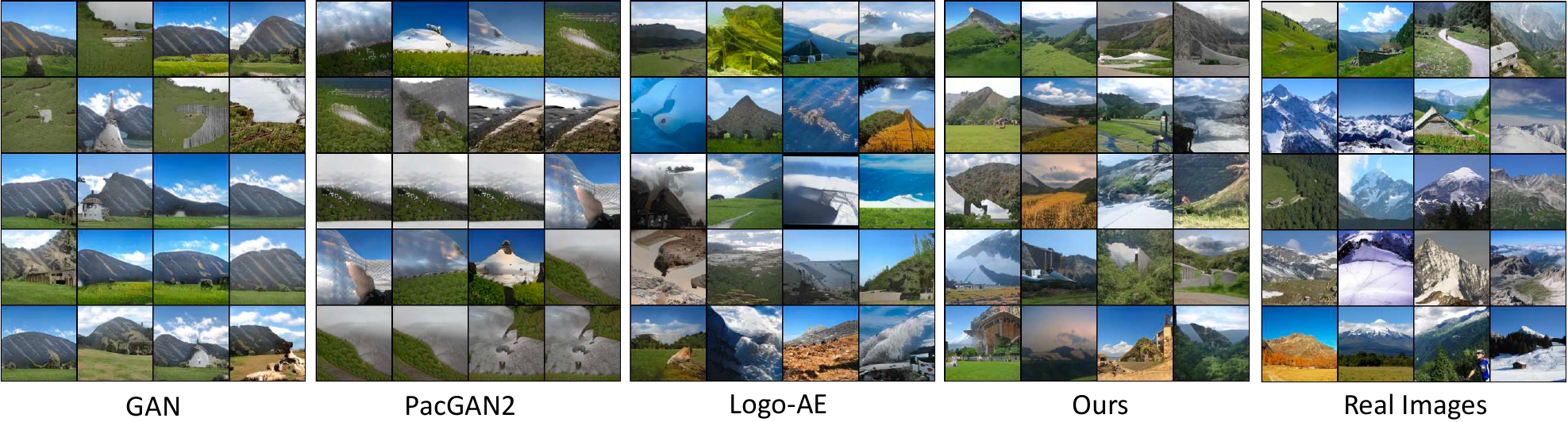}
\vspace{-17pt}
\caption{ImageNet samples with high classifier scores on the ``\texttt{alp}'' category.}
\end{subfigure}
\vspace{-19pt}
\caption{ImageNet~\cite{deng2009imagenet} samples from an unconditional GAN, PacGAN2~\cite{lin2018pacgan}, Logo-GAN AE~\cite{sage2018logo}, our self-conditioned GAN, and real images.  We observe that competing methods exhibit severe mode collapse, frequently producing similar images with a few patterns, whereas our method captures a wider range of diversity.  Samples are all sorted in rank order of a classifier score for a single category.  Please refer to our \href{http://selfcondgan.csail.mit.edu/}{website} for additional examples. }%
\lblfig{supp-diversity-imagenet}
\end{figure*}

We present more qualitative comparisons regarding the sample diversity of our method compared to various baselines on Places365 and ImageNet. \reffig{supp-diversity-places} (Places365) and \reffig{supp-diversity-imagenet} (ImageNet) visualize the sample diversity of various methods, all of which do not use labels.

\begin{figure*}[ht]
  \centering
\vspace{-10pt}
  \begin{subfigure}[b]{\textwidth}
    \includegraphics[width=\textwidth]{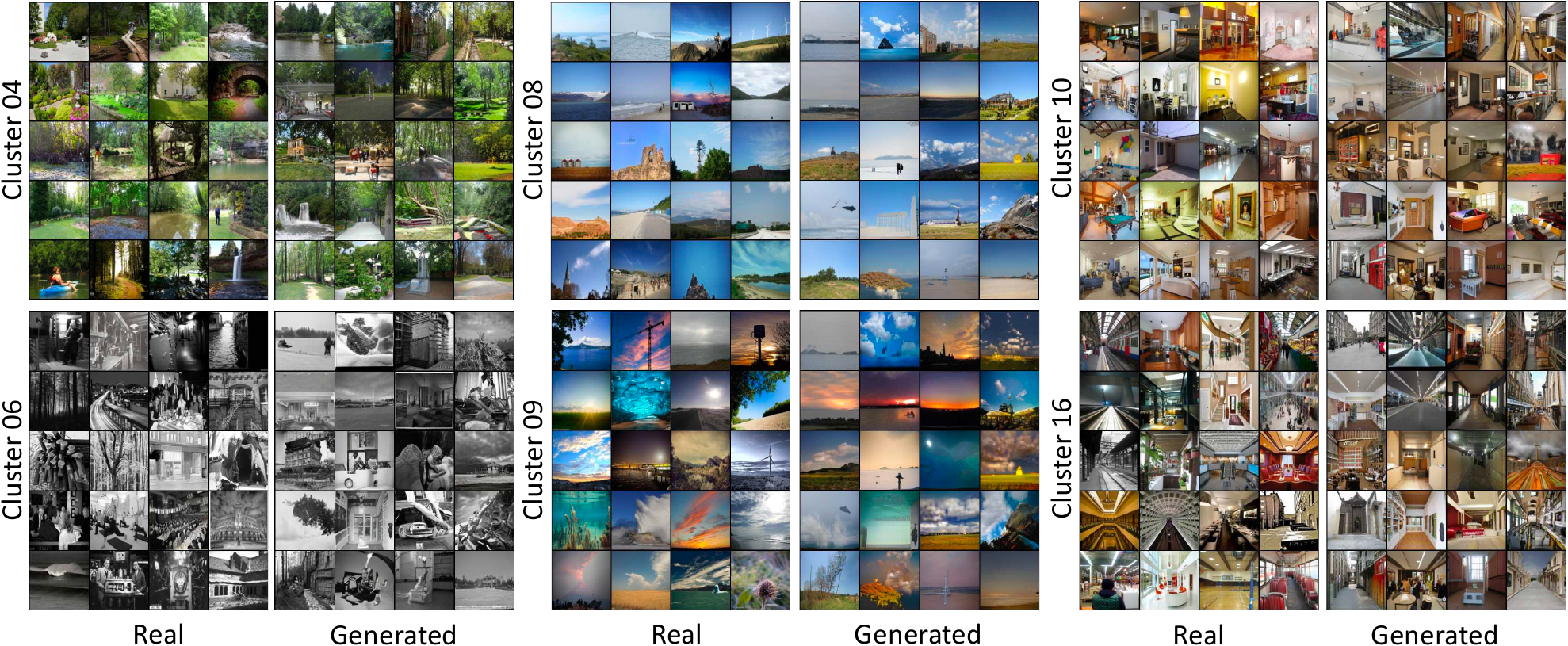}
    \caption{Inferred clusters and corresponding samples on the Places365 dataset.}\lblfig{supp-additional-places}
  \end{subfigure}\vspace{0.01cm}
  \begin{subfigure}[b]{\textwidth}
    \includegraphics[width=\textwidth]{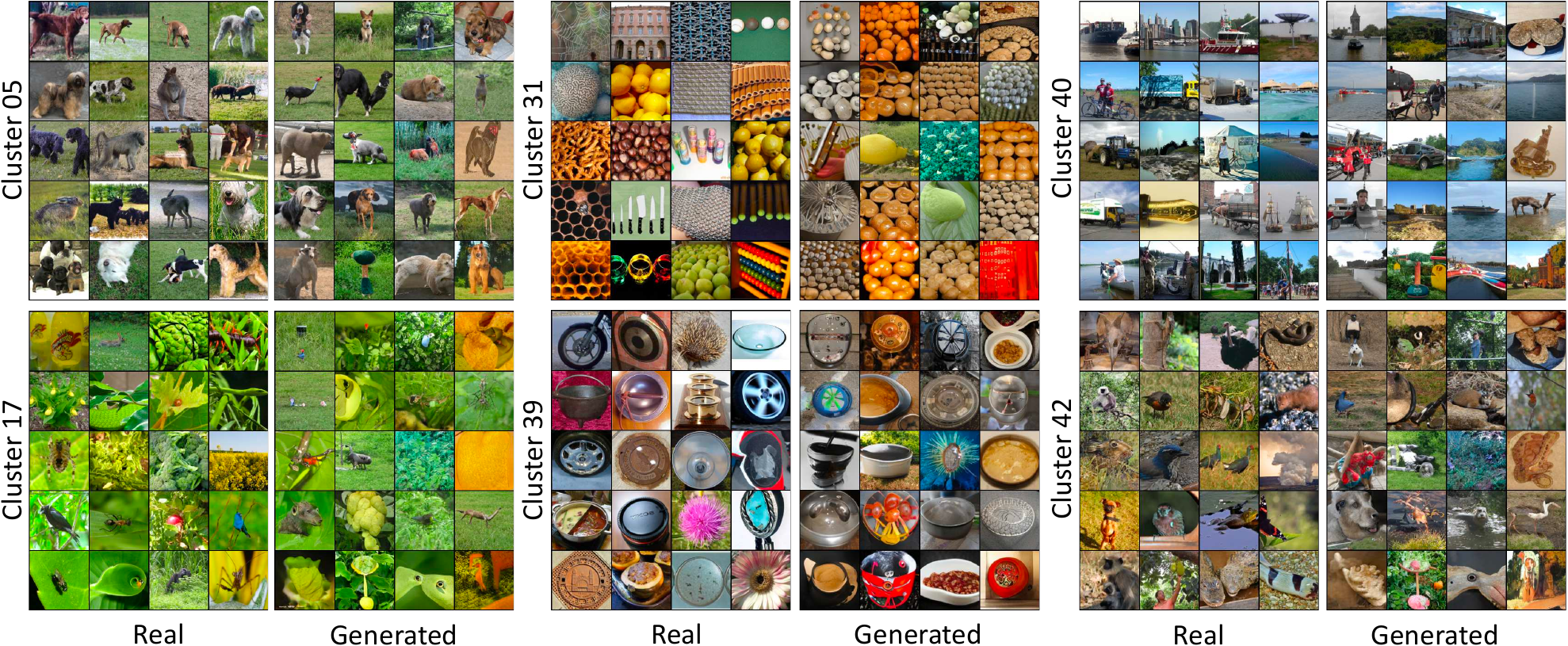}
    \caption{Inferred clusters and corresponding samples on the ImageNet dataset.}\lblfig{supp-additional-imagenet}
  \end{subfigure}
  \vspace{-18pt}
  \caption{Additional inferred clusters and generated samples for our method trained on ImageNet and Places365. Please refer to our \href{http://selfcondgan.csail.mit.edu/}{website} for additional examples. 
  }
  \lblfig{supp-additional-examples}
  \vspace{-8pt}
\end{figure*}

We also present additional visual results on Places365 and ImageNet.  In \reffig{supp-additional-places} (Places365) and \reffig{supp-additional-imagenet} (ImageNet), we show clusters inferred by our method and corresponding generated samples. More examples can be found on our \href{http://selfcondgan.csail.mit.edu/}{website}.

\end{appendices}

\end{document}